\documentclass[10pt,journal,compsoc]{IEEEtran}
\usepackage{amsmath}
\usepackage{cite}
\usepackage{algorithmic}
\usepackage{algorithm}
\usepackage[pdftex]{graphicx}
\usepackage{booktabs}
\usepackage{multirow}
\usepackage{array}
\usepackage{tabularx}
\usepackage{subfigure}

\usepackage{color}

\begin{document}
\title{Learning Discriminative 3D Shape Representations by View Discerning Networks}
\author{Biao Leng, Cheng Zhang, Xiaocheng Zhou, Cheng Xu, Kai Xu*
\thanks{
Corresponding author: Kai Xu (kevin.kai.xu@gmail.com)

Biao Leng is with the School of Computer Science \& Engineering, Beihang University, Beijing 100191,
and Research Institute of Beihang University in Shenzhen, Shenzhen 518057,
P.R. China (e-mail: lengbiao@buaa.edu.cn).

Cheng Zhang is with the School of Computer Science, Carnegie Mellon University, Pittsburgh, PA 15213 (e-mail: chengz2@andrew.cmu.edu)

Xiaochen Zhou is with the School of Engineering and Applied Science, Washington University in St.Louis, St. Louis, Missouri, MO 63130 (e-mail: zhouxiaochen@wustl.edu)

Cheng Xu is with the School of Computer Science \& Engineering, Beihang University, Beijing 100191,
P.R. China (e-mail: cxu@buaa.edu.cn).

Kai Xu is with the School of Computer, National University of Defense Technology, Changsha 410073,
P.R. China (e-mail: kevin.kai.xu@gmail.com).
}}
\maketitle
\begin{abstract}
In view-based 3D shape recognition, extracting discriminative visual representation of 3D shapes from projected images is considered the core problem. Projections with low discriminative ability can adversely influence the final 3D shape representation. Especially under the real situations with background clutter and object occlusion, the adverse effect is even more severe. To resolve this problem, we propose a novel deep neural network, View Discerning Network, which learns to judge the quality of views and adjust their contributions to the representation of shapes. In this network, a Score Generation Unit is devised to evaluate the quality of each projected image with score vectors. These score vectors are used to weight the  image features and the weighted features perform much better than original features in 3D shape recognition task. In particular, we introduce two structures of Score Generation Unit, Channel-wise Score Unit and Part-wise Score Unit, to assess the quality of feature maps from different perspectives. Our network aggregates features and scores in an end-to-end framework, so that final shape descriptors are directly obtained from its output. Our experiments on ModelNet and ShapeNet Core55 show that View Discerning Network outperforms the state-of-the-arts in terms of the retrieval task, with excellent robustness against background clutter and object occlusion.
\end{abstract}


\section{Introduction}
\IEEEPARstart{R}{ecent} advancements in 3D graphics technology have witnessed extensive application of 3D shapes in many areas, such as VR/AR, digital design, electronic entertainment and so on. As a result, a lot more 3D shapes are generated for multiple usages, which makes it possible, as well as crucial, to develop effective 3D shape recognition methods. Current approaches for 3D shape recognition can be generally classified into two categories: 3D model-based methods and view-based ones. 3D model-based methods like \cite{beneckart},\cite{matsuda},\cite{hilaga},\cite{jsun},\cite{Sinha},\cite{pointnet},\cite{pointnet++} extract high-level descriptors directly from the raw representation of 3D shapes. By contrast, view-based methods like \cite{ananliu},\cite{ananliu2},\cite{yanzhang},\cite{xiangbai},\cite{kaixu} aim to extract features from 2D images of a 3D shape. The final shape descriptor is constructed from these view features. Moreover, with the great success of deep learning techniques in visual recognition and classification tasks \cite{jkim},\cite{bianco},\cite{qingshanliu}, view-based methods employing Convolutional Neural Network (CNN) have achieved impressive performance and attracted much research attention.

Central to most existing view-based methods is the aggregation of features extracted from multiple images to form a compact and discriminative shape feature. Among these methods, two popular feature fusion paradigms of visual features \cite{hangsu},\cite{Edward} have achieved state-of-the-art performance on various 3D shape dataset. MVCNN\cite{hangsu} adopts max-pooling operation to fuse multi-view features
extracted by VGG-M \cite{chatfield}. On the other hand, \cite{Edward} presents a novel concatenation technique with the multi-view images. However, certain images with low discriminative ability can cause a negative effect on shape features. Furthermore, the adverse influence is even more severe under real scenes with background clutter and object occlusion, which poses challenges to the application of 3D shape recognition approaches.

\begin{figure}
\centering
\subfigure[]{
\includegraphics[width = \linewidth]{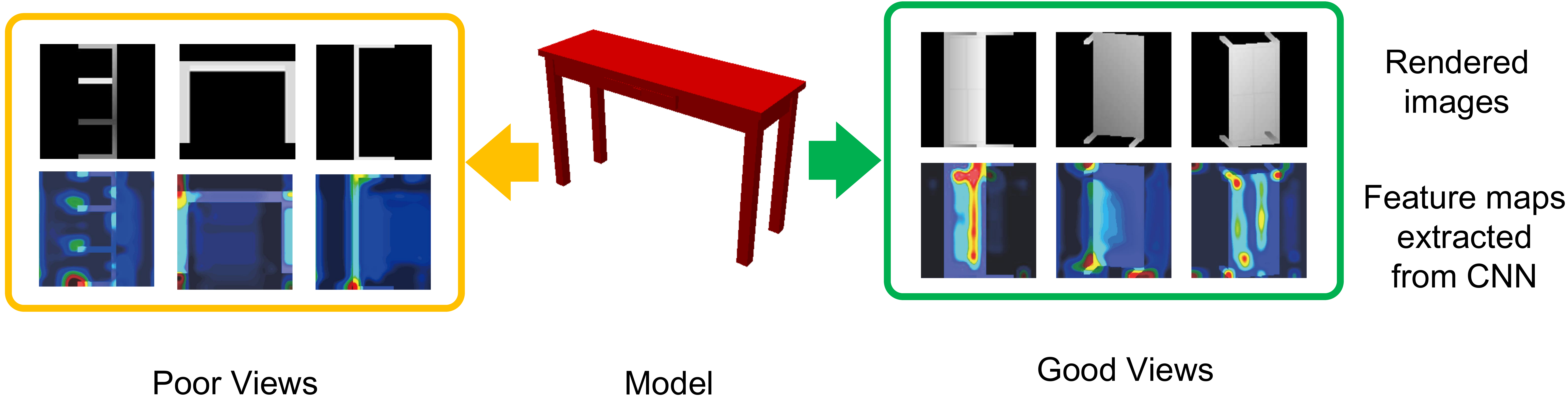}
}
\subfigure[]{
\includegraphics[width = \linewidth]{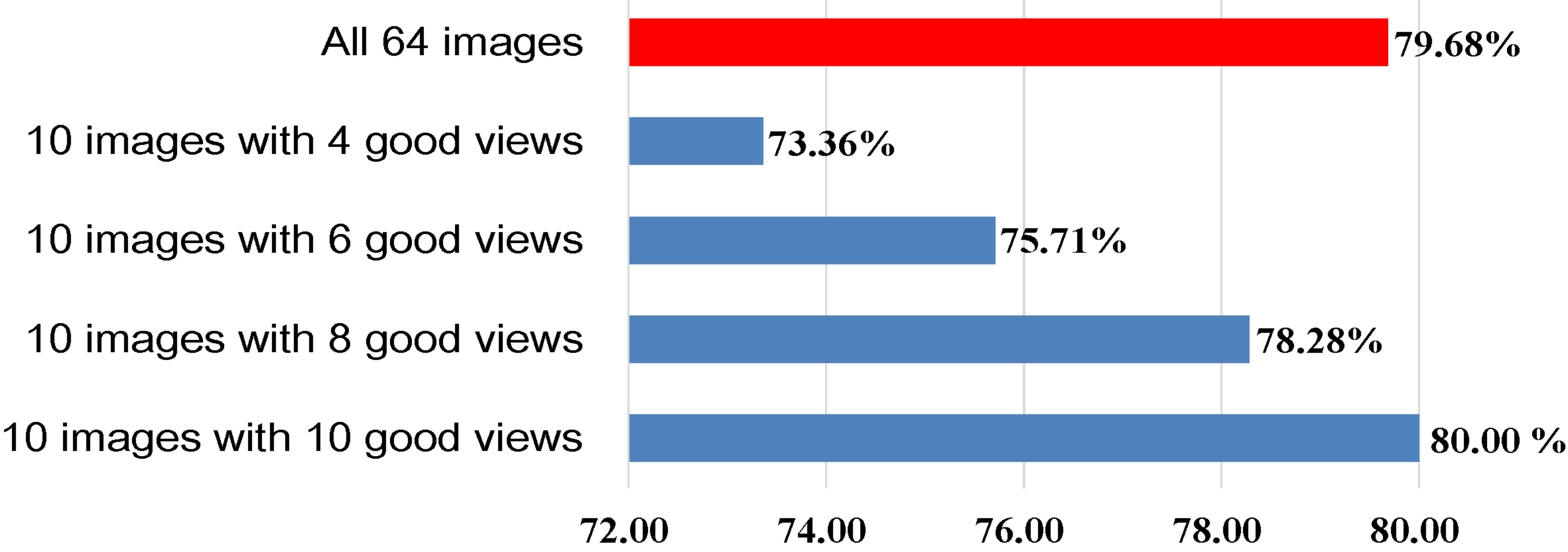}
}
\caption{(a) Examples of good views and poor views. For each view projection, the corresponding feature map is presented, representing the neuron activations of CNN, where CNN cannot learn discriminative information from poor views, as reflected by the relative low activation values. Good views activate more prominently, which is beneficial for shape recognition.
(b) An analysis of view quality on 3D shape recognition task.
Red bar: Recognition performance obtained by using uniformly sampled $64$ views.
Blue bars: Performance with only $10$ images, with increasing proportion of good views.
Using $10$ good views achieves even better performance than $64$ uniformly sampled views while saving a lot of computation. This strongly motivated us to develop the view discerning network for highly effective shape representation.}
\label{goodview}
\end{figure}

The above issue motivated us to study how the quality of different views affects the recognition results. The experiment shown in Fig. \ref{goodview} demonstrates the influence of view quality on shape recognition.
We heuristically classify the views of a 3D shape into two categories: poor views and good views, through setting
a threshold on the average activation values of feature maps extracted by CNN.
More details about the experiment can be found in Section~\ref{sec:influence}.
The results show that using $10$ good views achieves even better performance than $64$ uniformly sampled views, suggesting that different views have significantly different influence on multi-view 3D shape representation. The 10 views lead to not only highly effective shape representation but also low computational cost, compared to computing all 64 rendered view features.
In a real 3D scene, background clutter and object occlusion could intensify the influence of different views. Although a generic network like MVCNN may be able to learn to discriminate input views based on their quality in a data-driven fashion, we find that the performance of max pooling operation (CNN+MAX) can be adversely affected by object occlusion and background clutter, which is shown in Fig. \ref{noisyDatasetsCurve}. It indicates that this popular maximum aggregation operation on multi-view 3D shape recognition is sensitive to local noise in images, because the value of the noise may be the local maximum. Instead of using maximum aggregation, we try to design a view-quality-aware network utilizing the deep understanding of CNN.
Our work aims to achieve view selection in a data-driven fashion, with an end-to-end trainable model producing
view-quality-aware shape representations directly. The model is able to discern the view quality and aggregating information from only discriminative and relevant views.

We propose an end-to-end 3D shape recognition model, named View Discerning Network, to assess the quality of views and aggregate their features based on the evaluation scores. Specifically, a Score Generation Unit is introduced to the standard CNN to evaluate the quality of each projected image with score vectors. These score vectors are then used to weigh the image features and their summation serves as the representation of the shape. The weighted features are further refined by CNN.
The network is trained with supervisions targeting to both shape classification, based on Softmax loss,
and shape class guided feature embedding, through minimizing a Contrastive loss.
With these supervision signals, views with high quality become dominant among others,
since they are assigned with higher scores.
In contrast, non-informative views will be deprecated.

In designing the Score Generation Unit, we develop two structures to focus on different aspects of image features. The first structure is called Channel-wise Score Unit. A channel corresponds to a filter of the CNN. And each filter produces a feature map for the image. For every channel, Channel-wise Score Unit generates a certain score which is applied to the feature map. Consequently, this structure emphasizes the difference of feature maps extracted by different filters. The second structure, named Part-wise Score Unit, concentrates on the difference of local region of the image. It measures each part of the feature map based on the regional information, yielding a score map with the same size of the feature map. And the feature maps from the same image all share the same score map.

The advantage of our proposed method lies in three aspects. First, the learned 3D shape representation via View Discerning Networks exhibits more discrimination on various 3D shape dataset. Second,  the quality of different views can be efficiently captured, making the aggregated representation robust to background clutter and object occlusion. Thirdly, the view discerning network is trained in an end-to-end manner and with
weak supervision of only object category information, thus easy to implement.

The contributions of this paper include:
\begin{itemize}
\item We propose a View Discerning Network to gauge the quality of projection views of 3D shapes, with weighted shape features for shape recognition task.

\item Two structures of Score Generation Unit, Channel-wise Score Unit and Part-wise Score Unit, are introduced to assess image quality from different perspectives.

\item On the retrieval task, our method outperforms other state-of-the-arts on ModelNet \cite{shapenets} and ShapeNet core55 \cite{shrecpaper}, and is shown to be robust against background clutter and object occlusion.
\end{itemize}

\section{Related Work}
To create efficient representations of 3D shapes for the recognition task, various kinds of shape descriptors have been proposed in past decades. Generally, they could be classified as model-based descriptors or view-based descriptors.
\begin{figure*}
\centering
\includegraphics[width=7in]{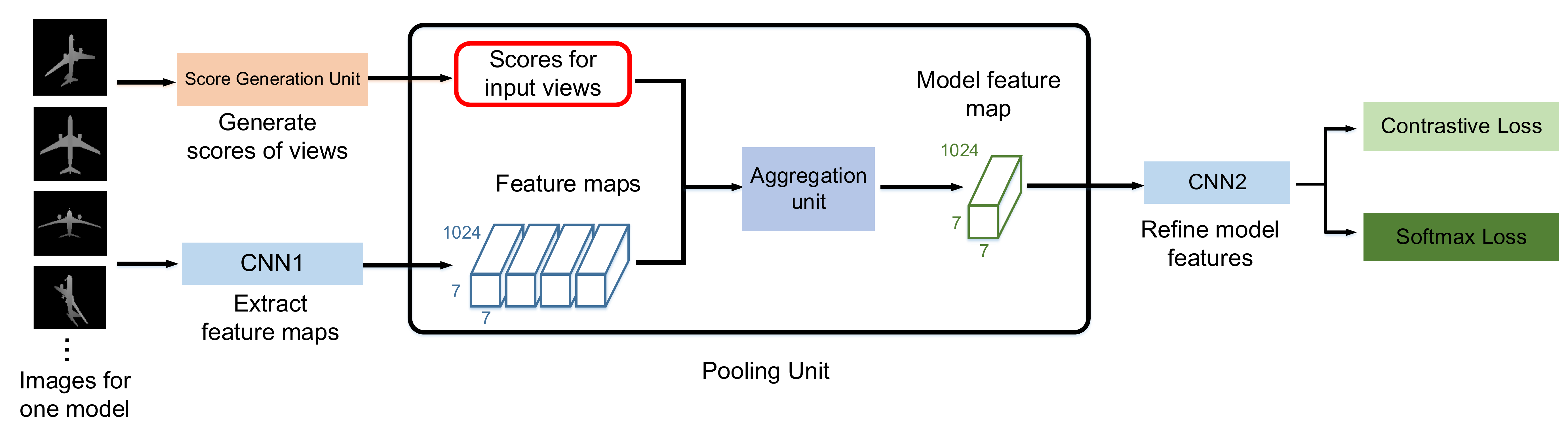}
\caption{Network architecture of View Discerning Network.}
\label{global}
\end{figure*}
Model-based methods extract shape features directly from raw 3D representations, such as point cloud \cite{beneckart},voxel grid \cite{matsuda},graph-based topological structures \cite{hilaga} and polygon mesh \cite{jsun}. In the past, many efforts have been done to improve the performance of model-based approaches. Common thinkings are mainly concerned with modifying the 3D representations or the feature extraction process. Li \emph{et al.} \cite{yangyanli} proposed field probing filters to cut down the high computational cost of 3D CNNs. They represented 3D shapes as volumetric fields and distributed probing points in the space. Their learning algorithm optimized the weights and locations of the points to sense 3D space intelligently. Bag-of-Words (Bows)\cite{Perona} model is widely used owing to the superiority performing as a natural image descriptor and reduction on computation complexity. Abdelrahman \emph{et al.} \cite{mostafa} proposed to include photometric information as a weight over the shape manifold of Weighted Heat Kernel Signature (W-HKS), thus enhancing the features extracted from shape information. Xie \emph{et al.} \cite{jinxie} modified traditional HKS methods to capture high-level features that are insensitive to deformations. They proposed a novel auto-encoder with the Fisher discrimination criterion to process multiscale data. Though past years witnessed obvious progress in model-based methods, they generally produce an inferior performance against view-based approaches\cite{qicharles}. It could be explained by two main drawbacks of raw 3D shape descriptors. First, the intrinsic structure flaws of naive 3D shape representation, such as occlusions and noise, impose bad influence on recognition results. Second, the dimension complexity and limited dataset make it not as fit as image descriptors for CNN, as cubically growing data heavily burdens the computation and tends to make classifiers overfitting, especially for databases with such a small scale (eg. ImageNet contains 3.2 million images \cite{jdeng} while ModelNet contains 150K shapes).

Utilizing view representations of 3D shapes, view-based methods achieve more satisfying recognition performance than model-based ones do. Despite that 2D images can't include complete information about a 3D shape, they have unignorable advantages, such as scalability and immunity to 3D noise. In this area, multi-view approaches are popular recently. For instance, Liu \emph{et al.} \cite{ananliu} proposed to enhance image representations by including information of 3D space. They constructed a view-graph model with spatial information of different views, thus transforming the issue of shape distance measurement into a graph matching problem. They \cite{ananliu2} also designed the multi-modal clique graph, which utilizes hyper-edges and edges to respectively link pairwise shape cliques and views within a clique. Such a graph design could strengthen inliers while suppressing outliers.

Two different kinds of feature descriptors are mainly used for multi-view shape recognition. The first one is hand-crafted descriptors, such as Zernike moments, SIFT, SURF and Fourier descriptor. They are designed according to characteristics of human vision and preserve distinguishing feature that our eyes are sensitive to. Zhang \emph{et al.} \cite{yanzhang} concatenated three different hand-crafted descriptors for each view to capture both contour and interior information of 3D objects. Bai \emph{et al.} \cite{xiangbai} encoded SIFT features of a pair of views into a single vector, which takes eigen-angle into consideration, hence enhancing 2D representation with spatial information. The second one is deep learning descriptors like the auto-encoder and CNN. They extract the feature from learning process rather than manually design. Zhu \emph{et al.}\cite{ZZhu} firstly exerted autoencoder for deep representation task on 3D shapes based on projected images. Recent advancement of deep neural network research has led to impressive achievements in many fields of computer vision, such as image super-resolution \cite{jkim}, recognition \cite{bianco},\cite{Bzhou} and classification \cite{qingshanliu}.  Also, many different modifications have been applied to CNNs for 3D shape recognition task. For instance, Shi \emph{et al.} \cite{baoguang} converted 3D shapes into panoramic views and did max-pooling in the CNN for each row of views. Bai \emph{et al.} \cite{bai} applied GPU acceleration to extracting view features and proposed an efficient context-based reranking algorithm. Guo \emph{et al.} \cite{haiyunguo} proposed a deep embedding network jointly supervised by triplet loss and classification loss and evaluated deep features extracted from different layers. Their network outperformed other State-of-the-Art methods a lot in SHREC'15, which demonstrates the effectiveness of deep learning features.

An important challenge for view-based methods is extracting distinguishing features of 2D representations. With the help of recent progress in deep neural network architecture \cite{szegedy},\cite{karen},\cite{szegedy2}, effective image features could be obtained through training. However, the robustness of shape feature is not only related to the CNN structure but also influenced by how these view features are aggregated into a compact one, which is another challenge in the area. Su \emph{et al.} \cite{hangsu} designed a novel CNN architecture, Multi-View CNN (MVCNN), to improve the synthesizing process by leveraging the learnability of deep neural networks. Instead of manually calculating the shape descriptors, the proposed MVCNN could learn to aggregate image features automatically. Nevertheless, different images of the same 3D shape contain different amounts of shape information. Though MVCNN performs well in recognizing 3D shapes through multi-view feature aggregation with an implicit feature weighting, our method achieves more efficient multi-view feature selection through an explicit, discriminative prediction of view quality. A desirable aggregation method ought to take the reliability of images into account. To cut down the influence of images with less discriminative shape information, we modify the traditional CNN by adding an additional branch to identify image quality. The network learns to adjust the weight of images based on the scores provided by the branch. Though Liu \emph{et al.} \cite{yuliu} proposed to predict a score for each image, they ignored the regional difference within a single image. And we designed two different networks to generate score vectors in a more detailed level. Moreover, inspired by \cite{haiyunguo}, we employ Contrastive loss at the end of the score branch to assist classification loss in the main part.

\section{The Influence of Quality of Images}
\label{sec:influence}
To analyze the influence of the image quality on the recognition performance, we make a relatively simple experiment on 3D shape retrieval. Fig. \ref{goodview} shows the analysis of view quality on 3D shape recognition. Mean Average Precision (MAP) is used as the evaluation metric.

\begin{figure*} [!h]
\centering
\includegraphics[height = 3in]{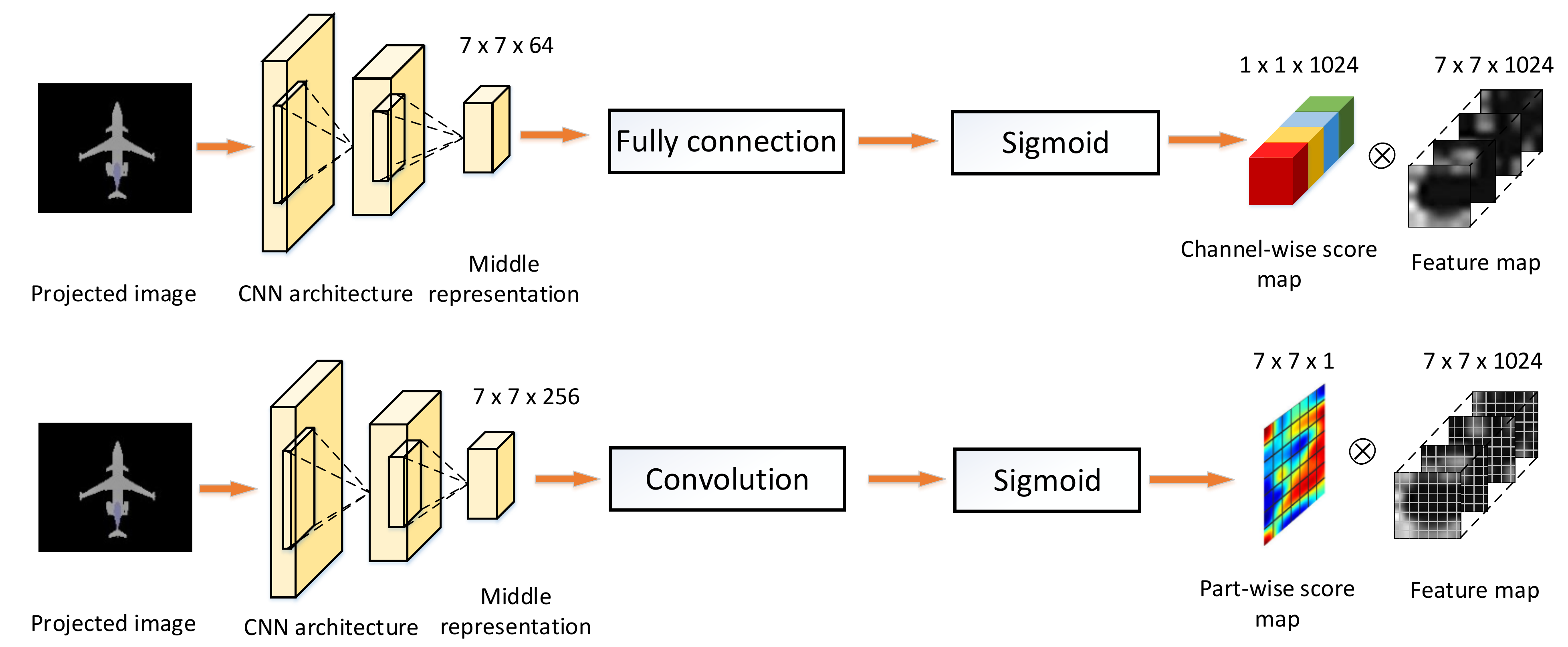}
\caption{Two structures of Score Generation Unit. Channel-wise Score Unit predicts a weight for each channel of the image feature. Part-wise Score Unit judges the regional quality of the feature map, where all the channels share the same score.}
\label{scorePic}
\end{figure*}

Firstly, we fetch 64 2D images for each shape from different angles for training a CNN\cite{szegedy}. The feature vector of each image is extracted from $pool5/7x7\_s1$ layer, which is then averaged as shape representation. These representations are utilized to calculate the cosine similarity as the shape similarity and then the MAP is computed as the result of \textbf{\emph{``All 64 images''}}, which is shown in Figure~\ref{goodview}.

Secondly, we utilize the cosine similarity among views to define the qualities of different images. For each image of a 3D shape, we utilize its visual feature from $pool5/7x7\_s1$ layer to calculate average cosine similarity with other 63 images. Generally, an efficient representation should achieve less intra-class distance and larger inter-class distance in retrieval task. Therefore, images with big average cosine similarity are closer to other images and we suppose these images can better represent the shape. Ranked by average cosine similarity, the top 10 images are defined as \textbf{\emph{good views}} and the lowest 10 images are defined as\textbf{\emph{ poor views}}. Some examples are given in Figure~\ref{goodview}. Good views and poor views are combined in different proportions. These mixtures are used for shape retrieval and computed as the results of  \textbf{\emph{``10 images with good views of different numbers''}}, which is shown in Figure~\ref{goodview}.

As is shown in Fig. \ref{goodview}, the collection of 10 images with a larger proportion of good views achieve better results. Therefore, it is confirmed that good views can lead to more effective shape representation. Besides, the image set containing only 10 good views performs even better than the image set with all the 64 views. Hence, the proportion of good views in the view collection indeed has an apparent influence on the performance of the network. To deal with the quality variance between the images, instead of view selection, we propose to take advantage of all image information by automatically controlling the weight of different view features. Judging the quality of the images with scores can offer more precise evaluation and cut down the influence of information deficiency.

\section{Overview}
We present a View Discerning Network for 3D shape recognition task, in which Score Generation Unit is proposed to judge the quality of input images, as is shown in the Fig. \ref{global}. We divide the structure of View Discerning Network into three parts. The first part is feature extraction. Different views rendered from 3D shapes is input into CNN1 where features are primitively extracted for the further work. The second part is Score Generation Unit whose input is the same as CNN1. It calculates the score distribution and then generates the score vector for each view. In the third part, score vectors and image features are aggregated in Pooling Unit to produce a weighted shape feature and the aggregated feature is refined in CNN2. The output of CNN2 serves as shape descriptor for recognition task.

\section{View Discerning Network}
The goal of View Discerning Network is to extract discriminative 3D shape descriptors from view images of them. In other words, for every 3D shape, a group of depth images rendered from different viewpoints is input into the network, while the shape descriptor $\mathbf{S}$ is directly obtained from the output of it. We call this image set $V=\{I_1,I_2,...,I_n\}$, where $I$ means the input image and $n$ is the total number of images for one shape. So, View Discerning Network aims to learn a mapping function $f(V)=\mathbf{S}$. By transforming a set of images into a point in the feature space, the similarity of different shapes can be easily measured with the cosine similarity between corresponding points. For two shape descriptors $\mathbf{S_1}$ and $\mathbf{S_2}$, the cosine similarity function $C(\mathbf{S_1},\mathbf{S_2})$ is defined as:
\begin{equation}
C(\mathbf{S_1},\mathbf{S_2})=\frac{\mathbf{S_1}\mathbf{S_2}}{\|\mathbf{S_1}\|\|\mathbf{S_2}\|}
\end{equation}
The larger the cosine similarity is, the more likely these shapes are from the same category.

\subsection{Network Architecture}
Via the deep neural network, we can get a high-level descriptor $\mathbf{D_{I_i}}$ for every image $I_i$. And our objective is to obtain the 3D shape descriptor $\mathbf{D}$, which is extracted from each $I_i$. Let us define the extraction method as $\mathcal{F}$. Then $\mathbf{D}$ can be denoted as
\begin{equation}
\label{equ:D}
\mathbf{D}=\mathcal{F}(\mathbf{D_{I_1},D_{I_2},...,D_{I_n}})
\end{equation}

Our focus is to improve the function $\mathcal{F}$ with the proposed network. Different from traditional approaches which simply average-pool or max-pool these image descriptors, we pay attention to the difference between discriminability of different views and give each view ($I_i$) an exclusive score $\mathbf{W_{I_i}}$ to adjust its influence on $\mathbf{D}$. Hence, in this paper, $\mathcal{F}$ can be defined as
\begin{equation}
\label{equ:weight}
\mathcal{F}(\mathbf{D_{I_1},D_{I_2},...,D_{I_n}})=\sum_{i=0}^n \mathbf{W_{I_i}\cdot D_{I_i}}
\end{equation}
where $\cdot$ denotes the Hadamard product. Furthermore, $\mathbf{D}$ in the function is not the final result, but an input to the rest of the network (CNN2), at the end of which we can get a more refined feature.

In this paper, View Discerning Network can be divided into three parts, CNN1 for feature extraction, an auxiliary network for scoring and CNN2 with Pooling Unit for aggregation and feature refinement. Fig. \ref{global} shows the structure of the View Discerning Network. For the first part, CNN1 is utilized to extract features from 2D images. For the second part, Score Generation Unit, an auxiliary network that generates scores for input images, is designed as a CNN which will be introduced in detail in the next section. Then in the third part, features from CNN1 and scores for views from Score Generation Unit are aggregated as weighted features in Pooling Unit and the weighted features are refined in CNN2 for further feature extraction.

In the structure, both CNN1 and Score Generation Unit receive the same images as input. Besides, CNN1 and CNN2 are obtained by dividing GoogLeNet with Batch Normalization into two parts. To be specific, CNN1 is parallel with Score Generation Unit and their output blobs converge at the Pooling Unit. The scores produced by Score Generation Unit and features extracted by CNN1 are of the same size so that we can get the weighted features through element-wise multiplication. But we do not intend to predict a weight on every dimension of the feature, as it is too complicated to learn. Instead, an original score vector is generated first, whose size is much smaller. Then we extend the size of it by concatenating its copies to produce the final score, which will be discussed detailedly later. Then, like MVCNN architecture introduced in \cite{hangsu}, the weighted 3D shape feature requires further learning to come out with a more discriminative feature. So, after merging score vectors and features together, we input the aggregated feature into CNN2 so that features passing through CNN2 are supervised by multiple loss layers.

As for the aggregation process in the Pooling Unit, the Hadamard product of the image features and scores are calculated as weighted features. Then, for weighted image features from each shape, we add up all of them as the weighted feature of the shape.

Apart from these, to improve the quality of the network, we adopt multi-loss training as our supervisory method. To speak in detail, we jointly employ Contrastive loss \cite{yisun} and Softmax loss. Softmax loss serves as the main supervisory signal, while Contrastive loss functions as an auxiliary signal.

\subsection{Score Generation Unit}
Score Generation Unit is aimed at judging the quality of images. Such a task is unpractical for humans as the dataset is too large to handle. Besides, people tend to evaluate images subjectively. So we let the deep neural network to do this scoring job.

Score Generation Unit is desired to generate score vectors to make a dimension-level judgment for the image features. To be specific, the shape of the image feature is [7 $\times$ 7 $\times$ 1024], where 7 $\times$ 7 refers to the size of a feature map. And 1024 filters in the network produce 1024 unique feature maps, forming the image feature. We design two structures for Score Generation Unit, Channel-wise Score Unit (CSU) and Part-wise Score Unit (PSU). The first method pays more attention to the quality difference between feature maps. And the second one focuses more on the regional quality of an image. Fig. \ref{scorePic} demonstrates both structures.

As shown in Fig. \ref{scorePic}, CSU cares about the difference of global features extracted by 1024 filters. And it produces a single score for each feature map. As for the detailed structure, the raw images first pass through 3 convolutional layers and a fully-connected layer, transforming into a vector whose shape is [1 $\times$ 1 $\times$ 1024]. Then, a sigmoid layer further processes the vector, generating the score vector. At last, to match the data shape of features generated by CNN1, we make 7 $\times$ 7 copies for each score vector and concatenate them into a larger vector with [7 $\times$ 7 $\times$ 1024] elements.

Different from CSU, PSU concentrates on local information of the images instead of global quality. For each image, it produces a score map with 7 $\times$ 7 dimensions which correspond to the 7 $\times$ 7 elements of a feature map. As 1024 feature maps are derived from one image, they all share the same score map. Similar to CSU, images go through 3 convolutional layers, while the output is sized as [14 $\times$ 14 $\times$ 96]. Then, we employ an extra convolutional layer with a 2 $\times$ 2 kernel, yielding a [7 $\times$ 7 $\times$ 1] vector. The score map is obtained after another sigmoid layer. Next, we make 1024 copies and combine them together, forming a score vector sized as [7 $\times$ 7 $\times$ 1024].

\subsection{Optimization}
In this paper, Softmax loss $L_S$ and Contrastive loss $L_C$ are jointly employed in View Discerning Network. Stochastic gradient descent is utilized to updating the network parameters. We assume that a mini-batch contains $M$ shape pairs. Every two consecutive shapes are regarded as a pair. Then $L_C$ is formulated as
\begin{equation}
\label{equ:contra}
L_C=\frac{1}{2M}\sum_{i=1}^M[s_i{E}_i+(1-s_i)max(\mathcal{M}-{E}_i,0)]
\end{equation}
where
\begin{equation}
\label{equ:dis}
{E}_i=\|\mathbf{N_{2i-1}}-\mathbf{N_{2i}}\|_2^2
\end{equation}
$\mathbf{N_{2i-1}}$ and $\mathbf{N_{2i}}$ are obtained via L2-normalization of the shape features $\mathbf{F_{2i-1}}$ and $\mathbf{F_{2i}}$, which compose a shape pair. $s$ provides similarity information between them. If they are from the same category, $s$ is set to 1, otherwise set to 0. $\mathcal{M}$ denotes the desired distance between shape features of different categories, which is manually adjusted based on specific cases. According to Equation \ref{equ:contra}, Contrastive loss tends to minimize intra-class distance while increasing inter-class distance. Thus it can assist the training based on Softmax loss which has little constraint on intra-class distance. To make our data fit for this formulation, we organize the input shapes in pairs.

By combining $L_S$ and $L_C$, we can get the overall objective function here:
\begin{equation}
\begin{aligned}
\label{equ:L}
L=\frac{1}{2M}\{\sum_{j=1}^{2M}L_{S_j}+\sum_{i=1}^M[s_i{E}_i+(1-s_i)max(\mathcal{M}-{E}_i,0)]\}
\end{aligned}
\end{equation}
Since we've got the objective function, we now focus on the gradient backward propagation formulation of Score Generation Unit. It is noted that we introduced two structures for Score Generation Unit, which yield score vectors with 2 different sizes. However, both of them are duplicated so that their shape matches the shape of the feature maps. Thus, the derivative of $L$ with respect to $\mathbf{W_{I_i}}$ for both structures are the same as below:
\begin{equation}
\label{equ:LWI}
\frac{\partial L}{\partial \mathbf{W_{I_i}}}=\frac{\partial L}{\partial \mathbf{D}}\cdot\frac{\partial \mathbf{D}}{\partial \mathbf{W_{I_i}}}
\end{equation}
According to Equation \ref{equ:D} and Equation \ref{equ:weight}, the derivative of $\mathbf{D}$ with respect to $\mathbf{W_{I_i}}$ is represented as below:
\begin{equation}
\label{equ:DW}
\frac{\partial D(d)}{\partial W_{I_i}(d)}=D_{I_i}(d)
\end{equation}
where $d$ is the dimension index. And $D(d)$ refers to the value of $\mathbf{D}$ on the $d-th$ dimension.
From Equation \ref{equ:LWI} and Equation \ref{equ:DW}, we have the derivative of $L$ with respect to $W_{I_i}(d)$:
\begin{equation}
\label{equ:LW}
\frac{\partial L}{\partial W_{I_i}(d)}=\frac{\partial L}{\partial D(d)}\cdot D_{I_i}(d)
\end{equation}
From Equation \ref{equ:LW}, we can easily find out that the learning process of the score map is directly influenced by the value of the feature maps. As $D_{I_i}(d)$ gets larger, the gradient of the corresponding regional score becomes larger, which means this region is more sensitive. Thus, an active region of the feature map may lead to a higher score. Conversely, the scores for less active regions may keep constant with lower values. As a result, such different sensitivities help the network to learn more efficiently, which is further proved in Section~\ref{sec:converge}.
\begin{algorithm}
\label{alg:1}
\begin{algorithmic}[1]
\REQUIRE ~~\\
training set $V$, initialized parameters, learning rate $\lambda(t)$, maximum number of iterations $T$;
\ENSURE ~~\\
trained network parameters
\STATE $t\leftarrow 0$;
\WHILE{$t<T$}
\STATE Calculate the Contrastive loss and Softmax loss through forward propagation;
\STATE Update the parameters of CNN2 and calculate $\frac{\partial L}{\partial D}$ through backward propagation;
\STATE Calculate $\frac{\partial L}{\partial W_{I_i}(d)}$ with Equation \ref{equ:LW};
\STATE Update the parameters of Score Generation Unit through backward propagation;
\STATE $t\leftarrow t+1$
\ENDWHILE
\end{algorithmic}
\caption{Learning Process of Score Generation Unit}
\end{algorithm}

During the training, the parameters of each layer are continually updated. However, we only focus on several specific layers in above discussion, as they are the distinctive point of View Discerning Network. With all the formulations we have derived, we form Algorithm 1 that illustrates the whole training process.

\section{Experiment}
\subsection{Datasets}
In our experiment, we exert Princeton Modelnet dataset and large-scale 3D shape retrieval datasets in SHape REtrieve Contest (SHREC) 2016 to train our model and test the capability of our method on 3D shape recognition. To present the recognition performance intuitively, we conduct 3D shape retrieval tasks on the above datasets.

\emph{1) ModelNet:}
this dataset is composed of 662 categories with 127,915 shapes from daily life. The core of the dataset is compiled to a list of the most common object categories in the world. ModelNet includes two subsets, ModelNet 40 and ModelNet 10. ModelNet 40 contains 12,311 shapes from 40 categories and ModelNet 10 possesses 4,899 shapes from 10 categories, both of which are utilized in our experiment. For these two datasets, we randomly choose at most 80 shapes from every category for training and 20 shapes for testing.

\emph{2) ShapeNet Core55:}
this dataset is introduced in SHape REtrieval Contest (SHREC), which concentrates on the scalability of 3D shape retrieval methods. Hence, we exert this large-scale dataset in our experiment, which contains about 51,190 3D shapes from 55 common categories, each subdivided into 204 subcategories. This dataset is divided into two parts, normal dataset and perturbed dataset. shapes in normal datasets are all aligned in one direction and shapes in perturbed datasets are randomly rotated. All the shapes are divided into three parts, 70\% used for training, 10\% for validation and 20\% for testing.  To test the robustness of View Discerning Network, we adopt both normal datasets and perturbed datasets for the experiment.

\subsection{Evaluation Criteria}
In this paper, five different evaluation metrics are used to present the retrieval performance of View Discerning Network for comparison with State-of-the-Art methods. The evaluation criteria are listed as below:

\emph{1) Mean Average Precision (MAP)} is the average precision which takes the retrieval rankings into account.

\emph{2) Precision and Recall Curve (PR Curve)} represents the precision and recall of the result.

\emph{3) Area Under Curve (AUC)} is the mean area under the precision-recall curves, which is used as the standard for the performance of retrieval tasks.

\emph{4) F-Measure (F)} evaluates both the recall and precision by calculating the harmonic mean of them.

\emph{5) Normalized Discounted Cumulative Gain (NDCG)} is a statistic that assigns more weight to correct results near the top of the ranked list and gives less weight to that near the end of the list, which is based on the fact that people are more likely to notice the top-ranked results than others.

These evaluation metrics are defined in \cite{shapenets},\cite{shrec16paper},\cite{jegou}. In this paper, we exert different metrics in different datasets. MAP and AUC are used in judging the performance of the structure on ModelNet 40 and ModelNet 10, and in ShapeNet Core55 datasets, we adopt MAP, F-measure, and NCDG as the standard of the judgment. Besides, Precision-Recall Curves are used to intuitionally show the comparison between View Discerning Network and other methods on the State-of-the-Art.

\subsection{Implementation Details}
For each 3D shape in the training and testing dataset, we render it with 10 cameras on the unit sphere centered at the shape. The position of each camera $i$ is represented by the azimuth $\theta^{az}_i$ and elevation $\theta^{el}_i$. For the first 8 cameras, we set that $\theta^{az}_i=i*45^\circ$, $\theta^{el}_i=0^\circ$. As for the rest 2 cameras, the azimuth is $0^\circ$, and the elevation is $90^\circ$ and $-90^\circ$ respectively. Every rendered 2D image consists of 224 $\times$ 224 pixels. The CNN is initialized with GoogLeNet with Batch Normalization model pre-trained on ImageNet\cite{jdeng}. In the training phase, all 10 images from one shape are fed to the proposed network as input. To fit the Contrastive loss used in our method, we organize the input images into pairs. And we need to make sure that both positive pairs and negative pairs exist in one mini-batch to take care of the intra-class and inter-class distance simultaneously. Through our experiment, it is suitable to set half of the shape pairs positive and the rest negative. Then, we obtain features from $inception\_5a/output$ layer and aggregate them with score vectors from Score Generation Unit. Features from $inception\_5b/output$ layer are pooled to [1 $\times$ 1 $\times$ 1024] as the final features.

In the experiment, the initial learning rate is set to 2e-3, which is halved when training loss meets the plateaus. The weight decay is set to 2e-4 and the momentum is set to 0.9. The weight of Softmax loss functioning as supervisory signals is set to 1 and Contrastive loss is utilized as an auxiliary supervisory signal where the loss weight is set as 1 and parameter Margin is set to 1.4. As for the input, each mini-batch contains 100 images rendered from 10 3D shapes. We conduct our experiment on GTX TITAN X with 12G memory, Intel(R) Xeon(R) CPU E5-2620 v4 and 128GB RAM. Caffe is exerted as the deep learning framework.

\subsection{Comparison on Different View Numbers}
\begin{table}[!h]
\renewcommand{\arraystretch}{1.3}
  \centering
  \caption{Comparison on the number of images for one shape}
  \label{number}
    \begin{tabular}{p{1.2cm}<{\centering} c c c c c}
    \hline
    Number&4&6&8&10&12\cr
    \midrule
    MAP&83.84\%&84.07\%&85.05\%&85.79\%&85.82\%\cr
    \midrule
    AUC&84.92\%&85.12\%&86.08\%&86.81\%&86.84\%\cr
    \bottomrule
    \end{tabular}
\end{table}
In our proposed approach, the number of images $N$ for one shape has an influence on the result of the retrieval task. If we increase $N$, which means more angles are selected to render 2D images, the network can receive more shape information, thus helping to make the extracted feature more robust and discriminative. However, more images in one shape can also greatly increase the training difficulty and time cost, as well as demanding more graphic memory. Hence, the number of views is considered not to be too large but able to ensure enough information for 3D shape retrieval. In order to find an ideal view number for our experiment, we carry out several comparative experiments on ModelNet 40 by varying the value of $n$.

The comparison is shown in Table \ref{number}, where AUC and MAP are utilized to judge the capability of the network. From Table \ref{number}, we can find that more images for a shape lead to better results. Besides, with the increasing of view number, although the performance of View Discerning Network improves, the increasing speed of the performance drops and more graphic memory is gradually used up as well. To be specific, the results of the 10-view method are quite close to the 12-view one, only $0.03\%$ lower than the latter one. Thus, we conclude that 10 images for one shape are enough for 3D shape retrieval task. And in our further experiment, we will exert 10 images from different angles to represent the 3D shapes.

\subsection{Comparison on Different Positions of Pooling Unit}
\begin{figure}[!h]
\centering
\includegraphics[width = 3.3in]{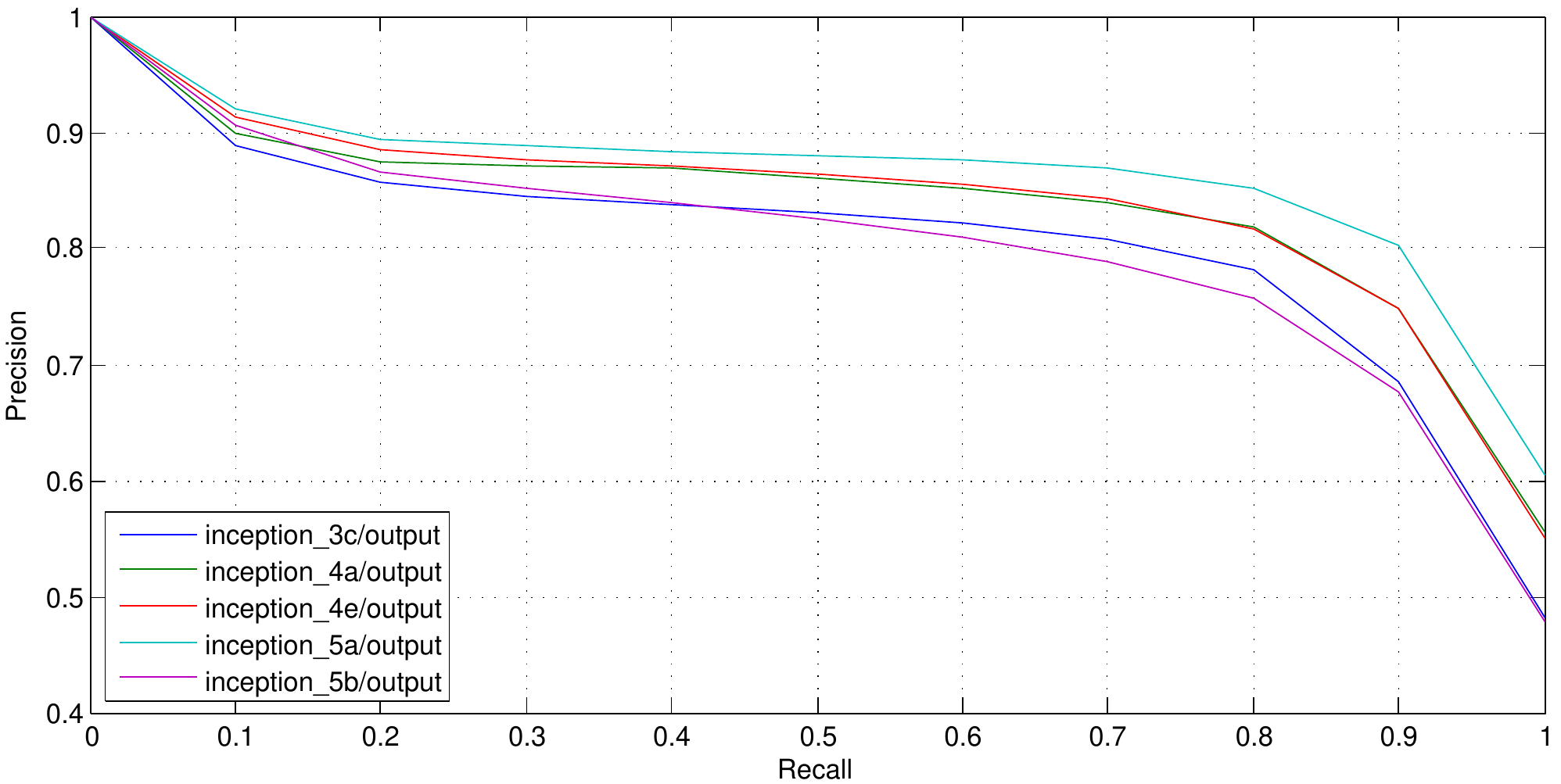}
\caption{Comparison of different positions of Pooling Unit.}
\label{position}
\end{figure}

In the design of View Discerning Network, it is necessary to find an ideal position of the Pooling Unit in the network, since it affects the recognition performance. For one thing, being near to the beginning of the network ensures sufficient information for CNN2, while the number of parameters will sharply increase and the aggregation of low-level abstract features will lead to the loss of discrimination. For another thing, if the Pooling Unit is set too close to the end of the network, features from input views will be highly abstract which indeed dismiss the useless information, while too many convolutional layers may result in lacking valid information. Apart from these, being too close to the end of network leaves less place for CNN2, which decrease the capability of further information extraction. Therefore, it is crucial to find a suitable position for feature aggregation. We do several experiments on the position of the Pooling Unit to optimize the performance of the network. Similar to the first comparison experiment, we utilize ModelNet 40 as the datasets and take AUC and MAP as evaluation metrics. The result of the experiment is shown in Fig. \ref{position}. Pooling unit is placed after the layers shown in the figure.

From the Fig. \ref{position}, shape features derived from Pooling Unit after $inception\_5a/output$ layer exhibits the best performance, which means Pooling Unit should be set right after $inception\_5a/output$ layer. Owing to that we ensure enough number of convolutional layers before the Pooling Unit, features of the original images have been sufficiently abstracted, during which indiscriminative information is dropped and abstracted features for aggregation remains. This position leads to a better performance. For one thing, performance on $inception\_5b/output$ layer is not satisfying. The reason may be that the output of $inception\_5b/output$ layer directly connects to the loss layers without convolutional layers for further feature extraction, for another word, there is no place left for CNN2. Compared with features from $inception\_5a/output$ layer, features from $inception\_5b/output$ layer lacks discriminative information. For another, other positions before $inception\_5a/output$ layer are less suitable as well. Though weighted features can be further processed by more convolutional layers, the insufficient abstraction for input views leads to decreasing quality of raw image features, which in the end shows in worse performance. To sum up, setting the Pooling Unit after the $inception\_5a/output$ layer can help the network to achieve the optimization.

\subsection{Comparison to State-of-the-Art Methods}

\begin{figure}
\centering
\subfigure[ModelNet 10]{
\includegraphics[width = 0.44\linewidth]{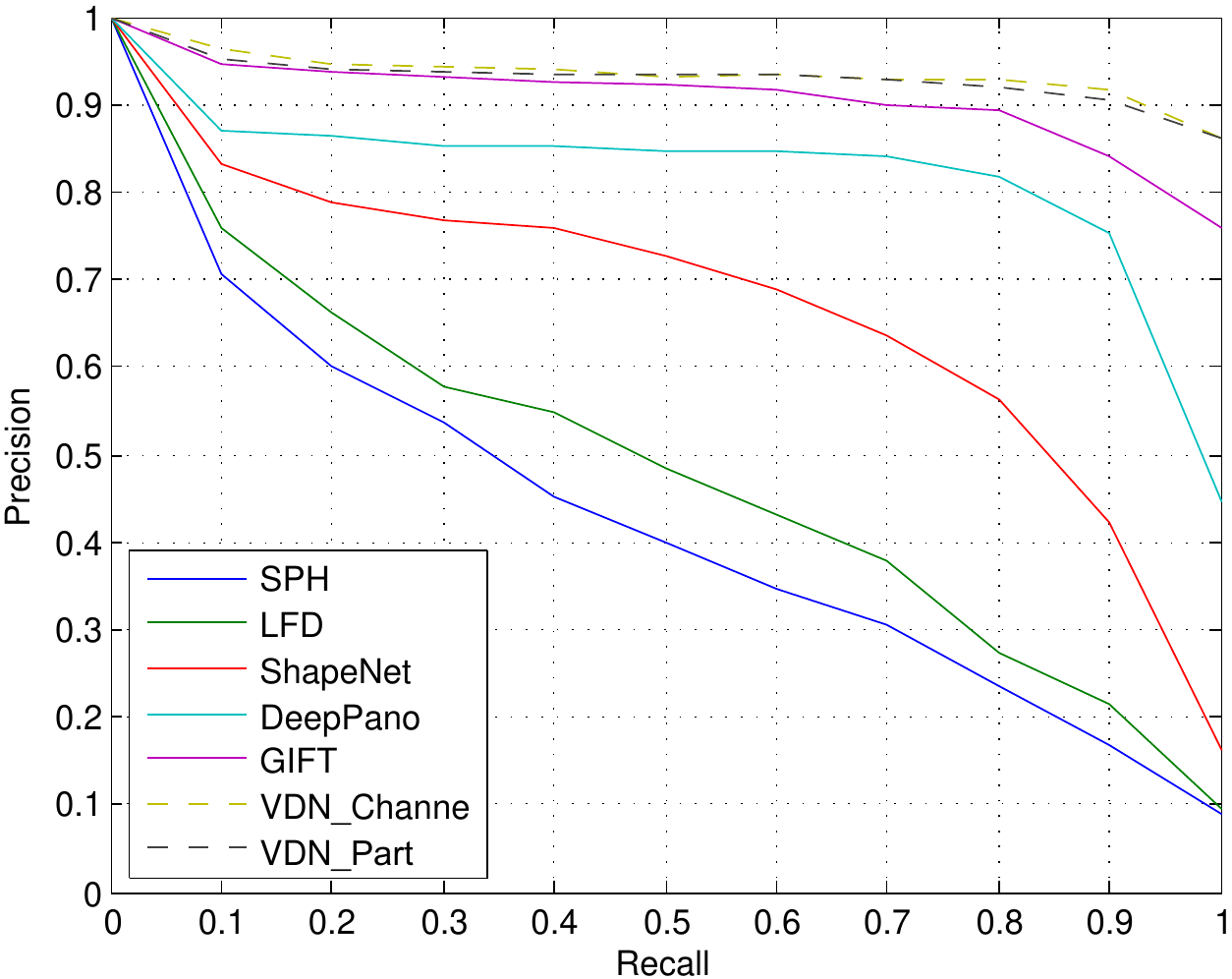}
\label{normalshrec}
}
\subfigure[ModelNet 40]{
\includegraphics[width = 0.44\linewidth]{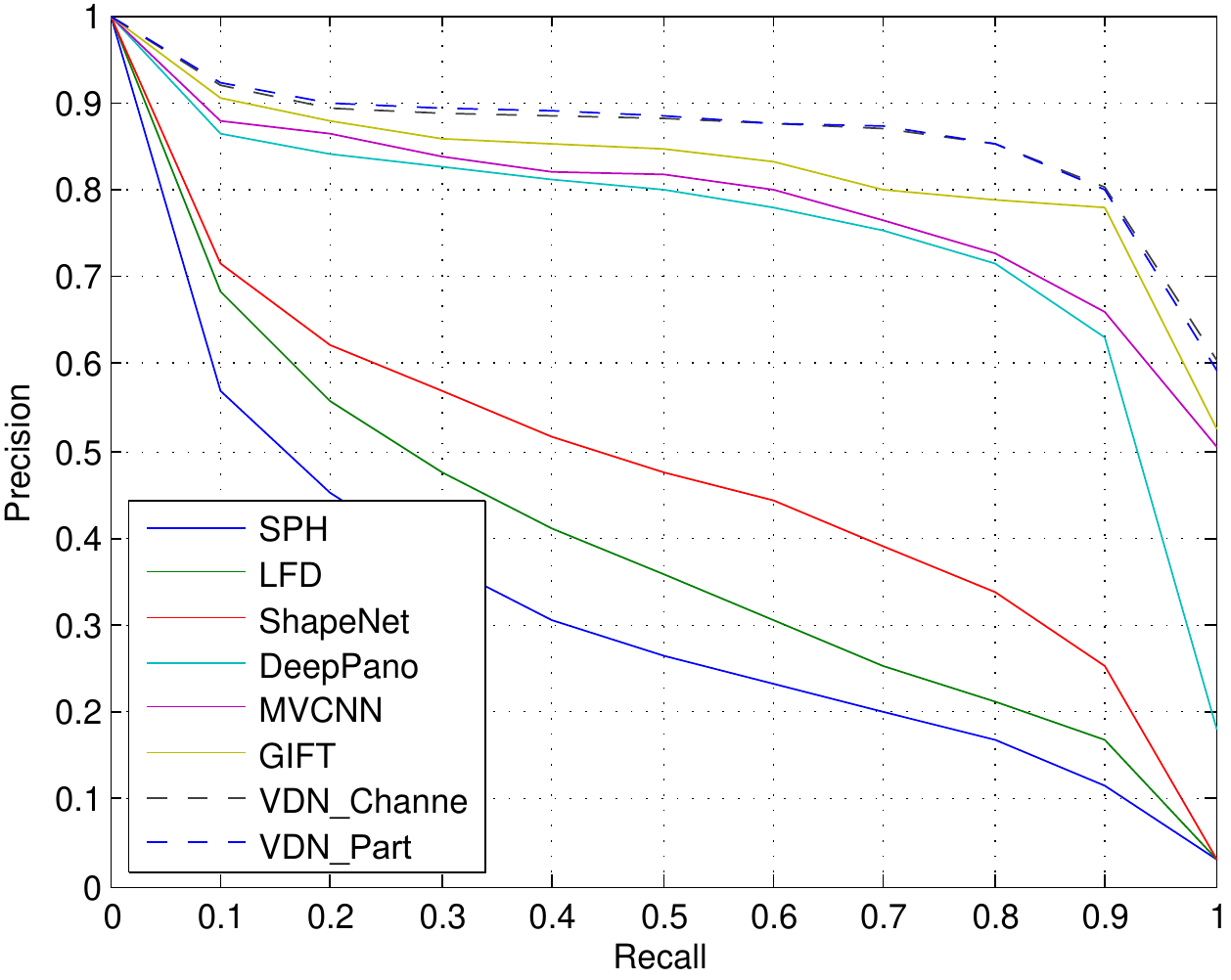}
\label{perturbedshrec}
}
\caption{PR Curve for ModelNet 10 and ModelNet 40.}
\label{modelnetcurve}
\end{figure}

\begin{table}[!h]
\renewcommand{\arraystretch}{1.3}
\centering
\caption{Performance comparision of 3D shape retrieval on ModelNet datasets}
\label{modelnet}
\begin{tabular}{lcccc}
\hline
\multirow{2}{*}{Method}&
\multicolumn{2}{c}{ModelNet 40}&\multicolumn{2}{c}{ModelNet 10}\cr
\cmidrule(lr){2-3} \cmidrule(lr){4-5}
&AUC&MAP &AUC&MAP\cr
\hline
SPH\cite{sph}&34.47\%&33.26\%&45.97\%&44.05\%\cr
LFD\cite{lfd}&42.04\%&40.91\%&51.70\%&49.82\%\cr
\hline
ShapeNets\cite{shapenets}&49.94\%&49.23\%&69.28\%&68.26\%\cr
DeepPano\cite{baoguang}&77.63\%&76.81\%&85.45\%&84.18\%\cr
MVCNN\cite{hangsu}&-&80.20\%&-&-\cr
GIFT\cite{bai}&83.10\%&81.94\%&92.35\%&91.12\%\cr
\hline
CNN\_MAX&83.14\%&81.90\%&91.75\%&91.31\%\cr
CNN\_AVE&83.07\%&81.85\%&91.66\%&91.23\%\cr
VDN\_MAX&83.81\%&82.49\%&-&-\cr
VDN\_Channel&87.45\%&86.46\%&93.57\%&93.24\%\cr
VDN\_Part&87.62\%&86.64\%&93.15\%&92.80\%\cr
\hline
\end{tabular}
\end{table}

\begin{table*}[!h]
\renewcommand{\arraystretch}{1.3}
\centering
\caption{Performance comparision of 3D shape retrieval on ShapeNet Core55 normal datasets}
\label{SHREC_normal}
\begin{tabular}{lccccccccc}
\hline
\multirow{2}{*}{Method}&
\multicolumn{3}{c}{Micro}&\multicolumn{3}{c}{Macro}&\multicolumn{3}{c}{Micro + Macro}\cr
\cmidrule(lr){2-10}
&F-measure&MAP&NCDG&F-measure&MAP&NCDG&F-measure&MAP&NCDG\cr
\hline
DB-FMCD-FUL-LCDR\cite{Tatsuma} &0.472&0.728&0.875&0.203&0.596&0.806&0.338&0.662&0.841\cr
CCMLT
&0.391&0.823&0.886&0.286&0.661&0.820&0.339&0.742&0.853\cr
ViewAggregation&0.582&0.829&0.904&0.201&0.711&0.846&0.392&0.770&0.875\cr
MVCNN\cite{hangsu}&0.764&0.873&0.899&0.575&0.817&0.880&0.670&0.845&0.890\cr
GIFT\cite{bai}&0.689&0.825&0.896&0.454&0.740&0.850&0.572&0.783&0.873\cr
\hline
VDN\_Channel&0.774&0.897&0.924&0.585&0.832&0.904&0.680&0.865&0.914\cr
VDN\_Part&0.776&0.902&0.923&0.588&0.841&0.909&0.682&0.872&0.916\cr
\hline
\end{tabular}
\end{table*}
\begin{table*}[!h]
\renewcommand{\arraystretch}{1.3}
\centering
\caption{Performance comparision of 3D shape retrieval on ShapeNet Core55 perturbed datasets}
\label{SHREC_perturbed}
\begin{tabular}{lccccccccc}
\hline
\multirow{2}{*}{Method}&
\multicolumn{3}{c}{Micro}&\multicolumn{3}{c}{Macro}&\multicolumn{3}{c}{Micro + Macro}\cr
\cmidrule(lr){2-10}
&F-measure&MAP&NCDG&F-measure&MAP&NCDG&F-measure&MAP&NCDG\cr
\hline
DB-FMCD-FUL-LCDR\cite{Tatsuma}
&0.413&0.638&0.838&0.166&0.493&0.743&0.290&0.566&0.791\cr
CCMLT
&0.246&0.600&0.776&0.163&0.478&0.695&0.205&0.539&0.736\cr
ViewAggregation&0.534&0.749&0.865&0.182&0.579&0.767&0.358&0.664&0.816\cr
MVCNN\cite{hangsu}&0.612&0.734&0.843&0.416&0.662&0.793&0.514&0.698&0.818\cr
GIFT\cite{bai}&0.661&0.811&0.889&0.423&0.730&0.843&0.542&0.770&0.866\cr
\hline
VDN\_Channel&0.679&0.844&0.900&0.439&0.741&0.854&0.559&0.793&0.877\cr
VDN\_Part&0.686&0.843&0.897&0.442&0.751&0.858&0.564&0.797&0.878\cr
\hline
\end{tabular}
\end{table*}
We compare our method with the State-of-the-Art in ModelNet and ShapeNet core55 datasets. In the ModelNet dataset, we contrast our method with several well-performed methods in both ModelNet 40 and ModelNet 10. And simultaneously we compare the result with the State-of-the-Art method in SHREC 16 in both normal dataset and perturbed dataset.

The retrieval result of the experiment on ModelNet is shown in Table \ref{modelnet} and we also make the comparison with the State-of-the-Art on the classification task of ModelNet, which is given in the supplementary materials. As can be seen in  Table \ref{modelnet}, the top two methods, SPH (Spherical Harmonic)\cite{sph} and LFD (Light Field descriptor) \cite{lfd}, utilized hand-crafted features, where Fourier descriptors and Zernike moments are taken into implementation. Then 3D ShapeNets\cite{shapenets}, DeepPano \cite{baoguang}, MVCNN \cite{hangsu} and GIFT \cite{bai} are methods based on deep learning techniques which present remarkable advantages over SPH and LFD. Besides, we compare our method with two baseline methods with different aggregation strategies. CNN\_MAX aggregates features by max-pooling, as is exerted in MVCNN, and CNN\_AVE adopt the average-pooling method. Both baselines are fine-tuned from GoogLeNet with Batch Normalization. Our two methods are named as VDN\_Channel and VDN\_Part, which means View Discerning Network with Channel-wise Score Unit and View Discerning Network with Part-wise Score Unit. In addition, VDN\_MAX aggregates the weighted features, multiplied by predicted scores, through the max-pooling operation as that in MVCNN.

As can be seen from Table \ref{modelnet}, compared with these methods, our methods show great competitiveness and advantages. From our perspective, the panoramic view in DeepPano does harm to the structure of the 3D shape and relatively enlarge the scale of input views as well, which limits the performance of this method. Besides, MVCNN proposes a classic strategy for multi-view retrieval. However, the max-pooling method in MVCNN is unable to utilize the complementary information from all images, which lower the experiment result. Then, the wrecking information of single view in GIFT hinders the capability of retrieval. Via Score Generation Unit, our methods ensure the discrimination of extracted features, leading to the outperformance. Specifically, compared with the best result in State-of-the-Art, in ModelNet 40, VDN\_Channel is $4.35\%$ higher in MAP and $4.52\%$ higher in AUC than GIFT. Besides, the max-pooling method in CNN\_MAX is unable to aggregate complementary information from different views, while CNN\_AVE can be seen as weak View Discerning Network setting all the score vector as 0.1, which is still insufficient for complementary information extraction. As is shown in Table \ref{modelnet}, in ModelNet 40, VDN\_Channel increases by $4.38\%$ in AUC and $4.61\%$ in  MAP than CNN\_AVE, in ModelNet 10, VDN\_Channel increases by $5.46\%$ in AUC and $5.88\%$ in MAP than CNN\_AVE. In addition, VDN\_MAX is $0.59\%$ higher in MAP than CNN\_MAX and it is shown that our explicitly score-based weighting mechanism is more efficient than the implicit maximum weighting as proposed in the MVCNN.

Fig. \ref{modelnetcurve} shows the Precision-Recall Curve of our methods and other State-of-the-Art methods, which intuitionally demonstrates the robustness and competitiveness of our methods.

In the experiment on ShapeNet core55 datasets, we adopt three new parameters, Macro-version (Macro), Micro-version (Micro) and mean of Macro-version and Micro-version (Macro+Micro), to judge the capability of the method. Macro gives an unweighted average over the whole datasets, while Micro is used to adjust the shape category sizes by giving a representative performance across categories. As for normalized discounted cumulative gain (NDCG), NDCG metric uses a graded relevance: 3 for perfect category and subcategory match in query and retrieval, 2 for category and subcategory both being same as the category, 1 for correct category and a sibling subcategory, and 0 for no match.

Table \ref{SHREC_normal} and Table \ref{SHREC_perturbed} show the comparison of View Discerning Network and the State-of-the-Art methods on normal datasets and perturbed datasets. The results of all the state-of-the-art methods are from the leaderboard of SHREC 2016 Large-scale 3D Shape Retrieval from ShapeNet Core55\cite{shrecresult}. In DB-FMCD-FUL-LCDR \cite{Tatsuma}, Feature Maps Covariance Descriptor (FMCD) is calculated on each depth-buffer image rendered for a given 3D shape and ranking scores is calculated by using the Locally Con-strained Diffusion Ranking (LCDR). Then in CCMLT, each 3D shape is rendered into 36 channel of data via concatenation of 36 2D projected images in sequence, where Multi-channel data is utilized to train a feature fusion matrix inside a CNN\cite{Krizhevsky}. As for ViewAggregation, CNN is exerted to extract features of rendered images and then these features are aggregated together. To avoid the disadvantages from consistent alignment, concatenating view-specific features in-order is adopted as the aggregation strategy.

From Table \ref{SHREC_normal} and Table \ref{SHREC_perturbed}, our methods outperform the results of State-of-the-Art methods on both normal dataset and perturbed dataset. The reason for the deficiency of DB-FMCD-FUL-LCDR method may be lacking the selecting of depth-butter images, where all of the rendered images are treated equally, leading to the lack of informative features for feature maps covariance descriptor. Besides, alignment method Point SVD\cite{alignment} is needed for the perturbed dataset, which takes up extra computing resource. Then, ViewAggregation may suffer from the consistent alignment on perturbed dataset as well, which increases the demand for computation. In fact, ShapeNet core55 datasets bring more challenges for 3D retrieval task. For one thing, 3D shapes with a larger scale are included in ShapeNet core55 datasets than ModelNet, making it difficult to obtain a satisfying retrieval result. For another, owing to the well-distributed split of 3D shapes and consistent alignment, shapes in ModelNet datasets are much cleaner than those in ShapeNet core55. Thus, perturbed datasets bring a further challenge for the network. While in View Discerning Network, both structure can to some extent deal with these challenges. Score Generation Unit possesses the ability to judge every single view, which means images from one shape are calculated independently. Consequently, our network performs better when facing this challenge. From Table  \ref{SHREC_normal} and Table \ref{SHREC_perturbed}, MVCNN shows better performance on normal datasets and GIFT does better on perturbed. While our methods outperform them on both datasets. In normal datasets, F-measure, MAP and NCDG in VDN\_Part are $1.2\%$, $2.7\%$ and $2.6\%$ higher than those in MVCNN on mean of Micro and Macro, and in perturbed datasets, F-measure, MAP and NCDG in VDN\_Channel is $2.2\%$, $2.7\%$ and $1.2\%$ better than those in GIFT. The comparison represents the scalability on large datasets and rotation robustness of our network.

\subsection{Experiments on Noisy Datasets}

\begin{figure}[h]
\centering
\includegraphics[width = 3.4in]{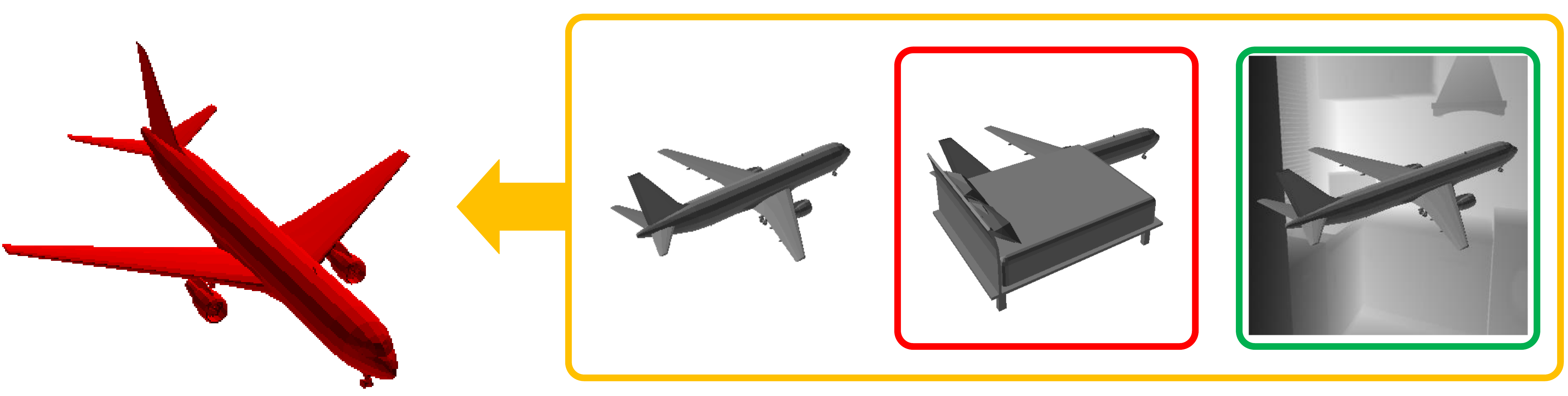}
\caption{Noise examples. In the yellow square, the left image is a normal view, the middle one shows the object occlusion and the right one is with a cluttered background.}
\label{noiseExamples}
\end{figure}

\begin{figure*}
\centering
\includegraphics[width = 7.5in]{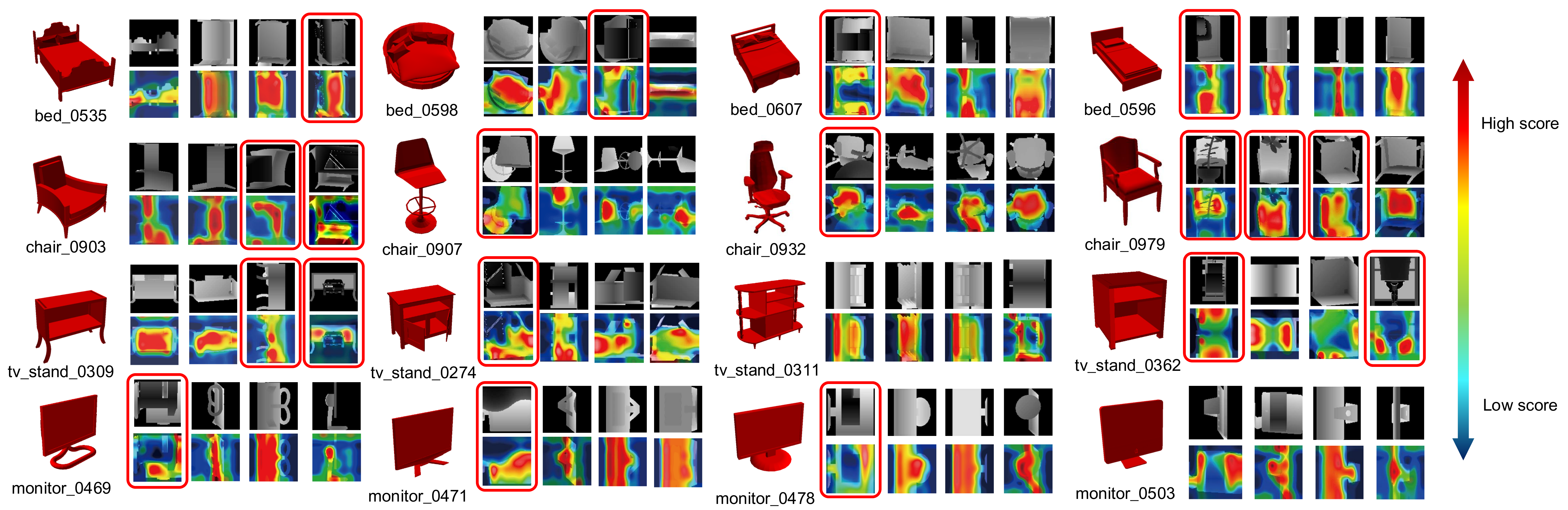}
\caption{The visualization of score maps. 16 examples of image sequences in 4 different categories are presented with heat maps. In each example, the original shape is in the left and the rendered images are followed, where rendered images are presented above and the visualized score maps are shown below. The images in red square represent the occlusion. According to the color bar on the right side, the red region represents a higher score and the green or blue region refers to a lower score.}
\label{list}
\end{figure*}

\begin{figure}[!h]
\centering
\subfigure[occlusion curve]{
\includegraphics[width = 0.44\linewidth]{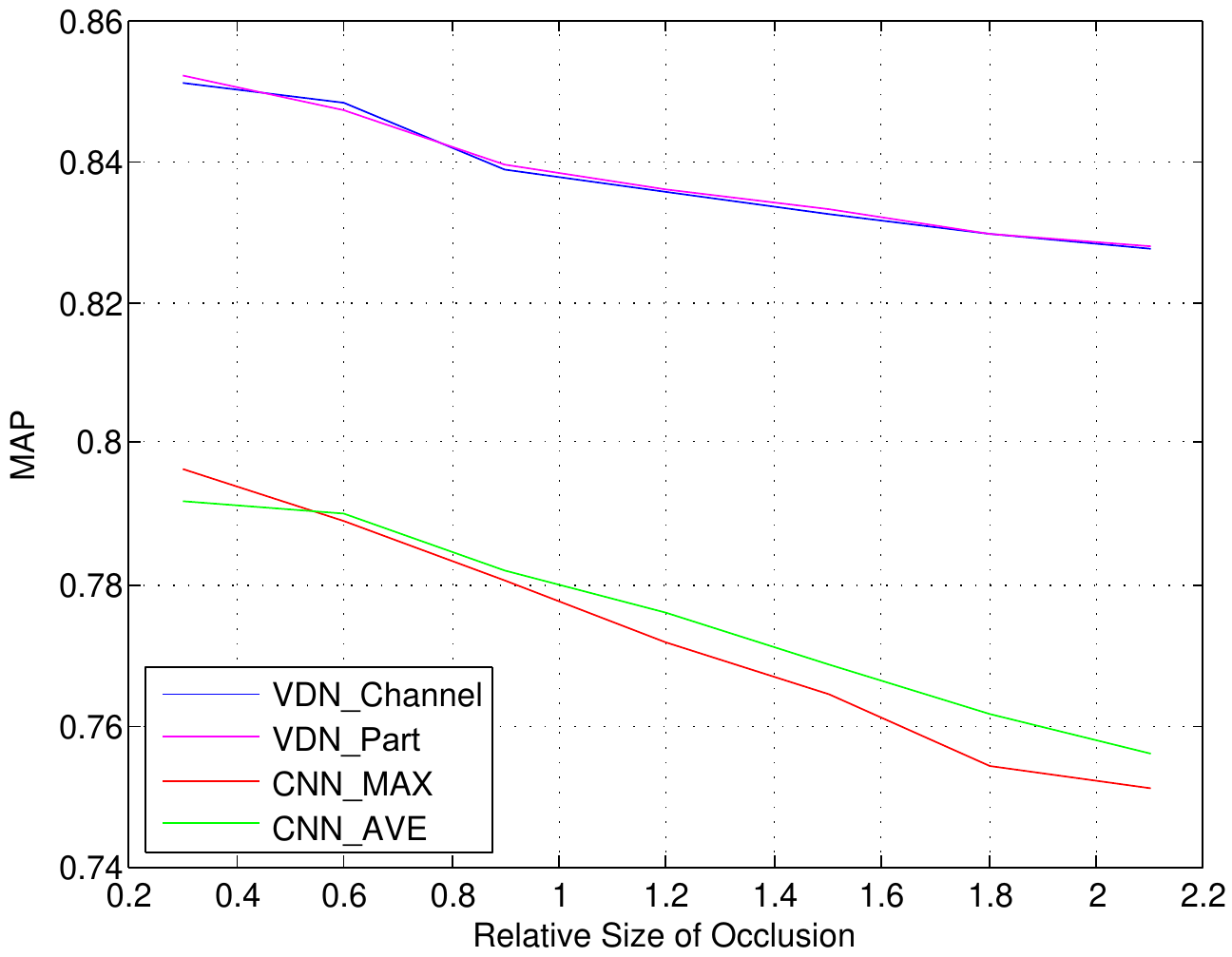}
\label{occlusionCurve}
}
\subfigure[background curve]{
\includegraphics[width = 0.44\linewidth]{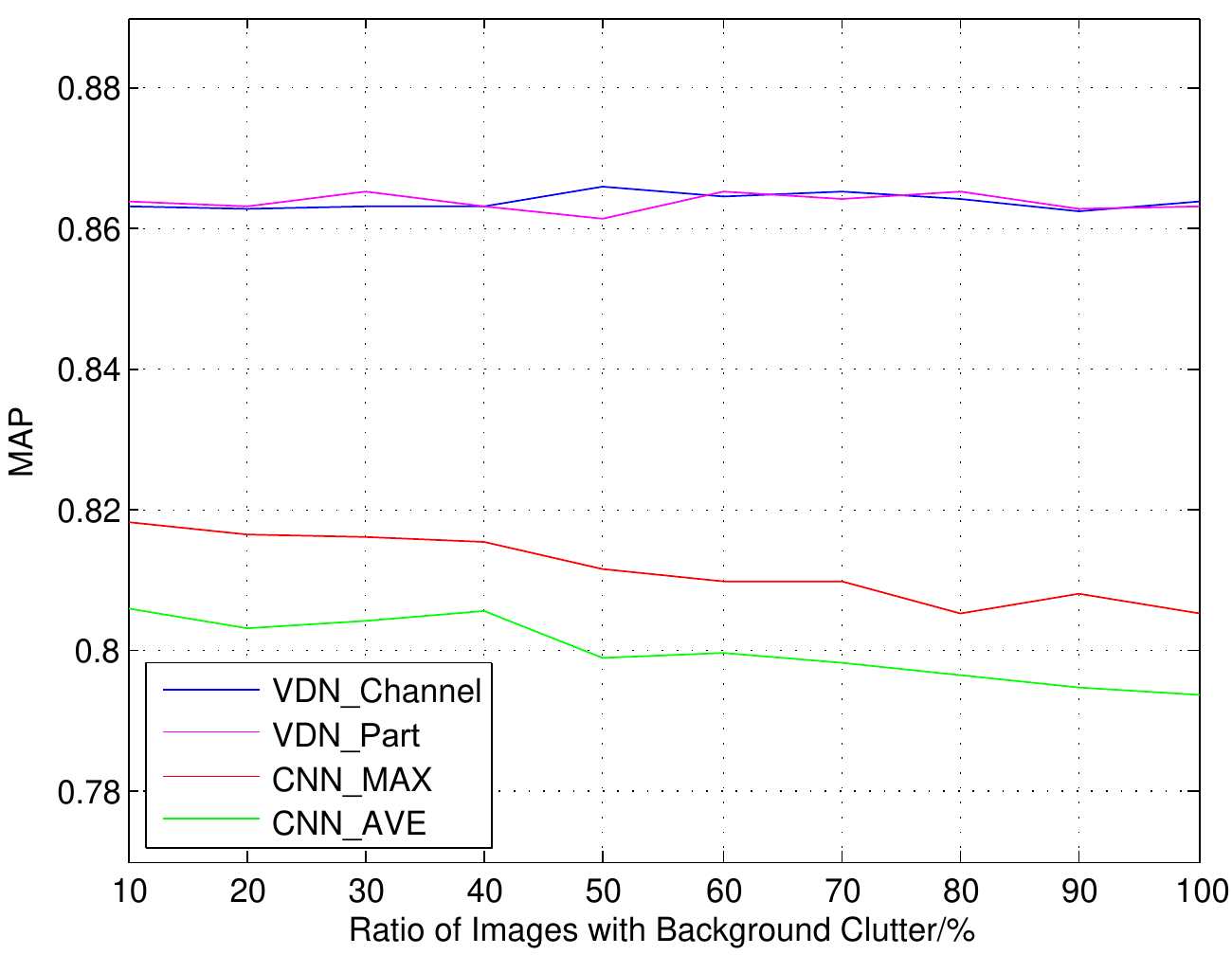}
\label{backgroundCurve}
}
\caption{Comparison of the performance on noisy datasets with baseline methods.}
\label{noisyDatasetsCurve}
\end{figure}

In the indoor 3D scene, the performance of view-based retrieval methods is limited by noise, mainly including object occlusion and background clutter. Fig.  \ref{noiseExamples} presents examples of the noise. In this experiment, we compare our proposed methods with baseline methods on the capability of resisting noise in a complex environment on ModelNet 40 dataset. The influence of object occlusion and background clutter are separately tested. For the object occlusion, we put an irrelevant 3D shape beside the target shape during the rendering process to produce noisy images. In the training phase, we set the size of the interfering shape as 1.2 times its original one, and in the testing phase, the relative size of occlusion varies from 0.3 to 2.1 for a thorough comparison. As for the background clutter, we randomly select a quarter of the images in each mini-batch to add a background during the training. Similarly, multiple experiments are conducted in the testing phase by altering the ratio of images with background clutter. The experiment results are presented in Fig. \ref{occlusionCurve} and Fig. \ref{backgroundCurve}.

As can be seen in Fig. \ref{occlusionCurve}, our methods outperform the baseline methods when facing the object occlusion. Besides, when the size of occlusion increases, the results of "CNN\_MAX" and "CNN\_AVE" drop faster than our methods. As the size of occlusion changes from 0.3 to 2.1, the MAPs of baseline methods reduce by around $4\%$. But our methods decrease only $2\%$. It is concluded that the baseline methods suffer apparent influence of the noise, which cannot be well handled via a simple max-pooling or average-pooling operation. In Fig. \ref{backgroundCurve}, the MAPs of baseline methods drop about $1.2\%$ when the proportion of noisy images increases from 10\% to 100\%, while our methods are almost not influenced by it. It is shown that our proposed method has the strong ability to discern this noisy information and then avoid its effect maximumly.

To present the validity of the scores, the score maps from Part-wise Scoring Network are visualized in Fig. \ref{list}. To make the results more intuitive, some rendered views are organized with occlusions. As the size of each score vector is 7 $\times$ 7, the scores are resized to 224 $\times$ 224 and presented as heat maps for a clear look. For every heat map, we cover it on the original image to show the region-wise corresponding relationship. As Fig. \ref{list} shows, the occluded parts in the images are covered by green or blue areas, which refer to low scores. In contrast, the red areas, which represent high scores, mostly cover the parts that are not occluded. Thus, we can conclude that the score maps successfully discriminate features from different regions. Although the score vector generated by Channel-wise Scoring Network is not visualized here, it also has such capability in evaluating images, which has been testified by its performance in previous experiments. Given the above results, it is confirmed that the Score Generation Unit owns the ability to avoid information loss and interference.

\begin{figure*}[!h]
\centering
\includegraphics[width = 7in]{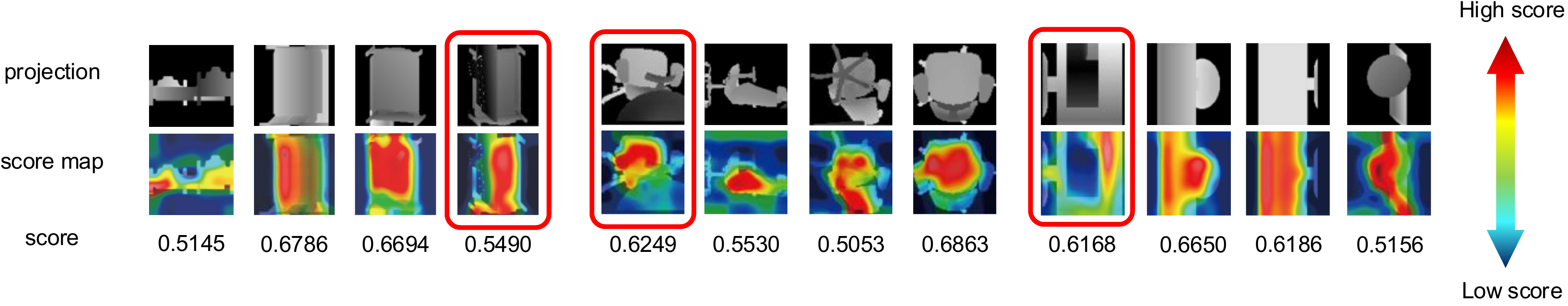}
\caption{Examples of images, score maps and corresponding single scores. We select 3 3D shapes from different categories for a clear demonstration. As is shown in the first row, 4 images are picked out for each 3D shape. The corresponding score maps lie in the second row. As for the last row, we put the single scores for these images here. The images surrounded by a red square is with occlusion. A color bar is presented on the right side, which suggests that the red means a higher score, while the blue means a lower score.}
\label{singlescore}
\end{figure*}

\subsection{Discussion on the Proposed Score Generation Unit}

\begin{table}[!h]
\renewcommand{\arraystretch}{1.4}
\centering
\caption{Performance comparison of retrieval results on Rough-level Score and Fine-level Score}
\label{SingleScoreTabel}
\scalebox{1.16}{
\begin{tabular}{lcccc}
\hline
\multirow{2}{*}{Method}&
\multicolumn{2}{c}{Normal ModelNet40}&\multicolumn{2}{c}{Occluded ModelNet40}\cr
\cmidrule(lr){2-3} \cmidrule(lr){4-5}
&AUC&MAP &AUC&MAP\cr
\hline
VDN\_Single&85.31\%&84.20\%&80.90\%&79.61\%\cr
VDN\_Channel&87.45\%&86.46\%&84.52\%&83.47\%\cr
VDN\_Part&87.62\%&86.64\%&84.55\%&83.51\%\cr
\hline
\end{tabular}}
\end{table}

\textbf{(a) single score v.s. proposed score map}

In order to evaluate the power of the proposed fine-level score map for images, we give a comprehensive comparison of the rough-level single score and the fine-level score map. As shown in Table \ref{SingleScoreTabel}, the method VDN\_Single is trained to directly predict a scalar weight for each view and these predicted global scores are used to weight image features. We did experiments both on the normal dataset and occluded dataset of ModelNet 40. In the normal ModelNet40 dataset,   the MAP of VDN\_Part is $2.31\%$ higher than VDN\_Single. Meanwhile, this gap is more obvious in the occluded dataset and the MAP of the region-wise score outperforms the single score by $3.65\%$. To further present the superiority of the fine-level score map, we also extracted the single score of each image from the VDN\_Single and compare them with the score maps from the VDN\_Part, which is shown in Fig. \ref{singlescore}. As we can see, the predicted single scores of certain occluded images are higher than other normal images. It is shown that the rough-level single score is difficult to give an accurate judgment of the occlusion and our fine-level score map can offer a more precise evaluation of the local occlusion of images. Therefore, the fine-level score map is more robust to background clutter and object occlusion.

\begin{table}[!h]
\renewcommand{\arraystretch}{1.4}
\centering
\caption{Comparison of retrieval results on Contrastive Loss}
\label{ContrastiveLossTable}

\begin{tabular}{lcccc}
\hline
\multirow{2}{*}{Method}&
\multicolumn{2}{c}{Softmax loss}&\multicolumn{2}{c}{+ Contrastive loss}\\
\cmidrule(lr){2-3} \cmidrule(lr){4-5} &AUC&MAP &AUC&MAP\\
\hline
CNN&79.67\%&78.42\%&80.04\%&78.80\%\\
VDN\_Channel&86.33\%&85.24\%&87.45\%&86.46\%\\
VDN\_Part&86.57\%&85.64\%&87.62\%&86.64\%\\
\hline
\end{tabular}
\end{table}

\textbf{(b) the effect of Contrastive loss on score generation unit}

\begin{figure}[h]
\centering
\includegraphics[width = 3.5in]{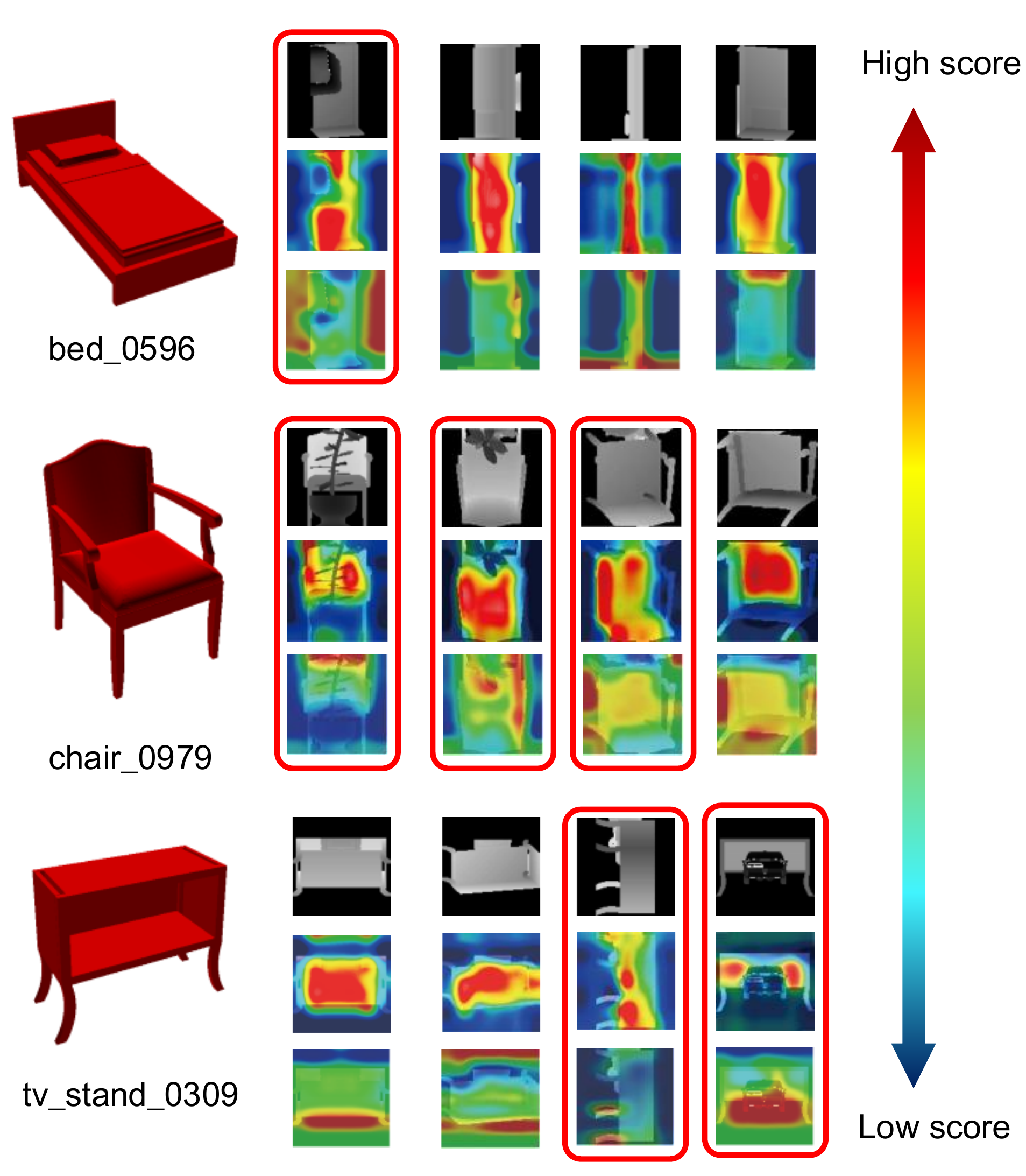}
\caption{Examples of images and corresponding score maps with and without Contrastive loss. Three 3D shapes are picked out for a clear demonstration. For each shape, in the first row, 4 images are selected. The corresponding score maps with Contrastive loss lie in the second row. And the last row shows the score maps without Contrastive loss. Also, we point out the occlusion with a red square and put a color bar is on the right side, which suggests that the red means a higher score, while the blue means a lower score.}
\label{contrastive}
\end{figure}
Compared with state-of-the-art MVCNN and GIFT which are supervised by softmax loss, the proposed view discerning network utilizes an extra Contrastive loss to guide feature learning and score generation. We evaluate the performance of different methods with and without Contrastive loss on ModelNet40 to study the effect of Contrastive loss on score generation unit. We set a baseline method CNN, which denotes that all image features are extracted from a standard CNN and then are average pooled to a feature vector as the final 3D shape representation. The retrieval results are shown in Table \ref{ContrastiveLossTable}. As we can see, the Contrastive loss can bring $0.38\%$ improvement on MAP for a standard CNN compared against CNN without Contrastive loss. However, VDN\_Channel receives a higher improvement of $1.12\%$ on MAP with Contrastive loss. The score maps from VDN\_Part with and without Contrastive loss are visualized in Fig. \ref{contrastive}. With help of Contrastive loss, the generated score map can be more effective in discriminating input image features. Therefore, the gain from Contrastive loss for our method is mainly due to the more effective generated score maps. And it indicates that Contrastive loss can enable the proposed score generation unit to distinguish local information more effectively because the generated score map is trained to make weighted features more optimized in the feature space.

\subsection{Comparison on Converging Rate of Loss}
\label{sec:converge}
\begin{figure}[!h]
\centering
\includegraphics[width = 3.3in]{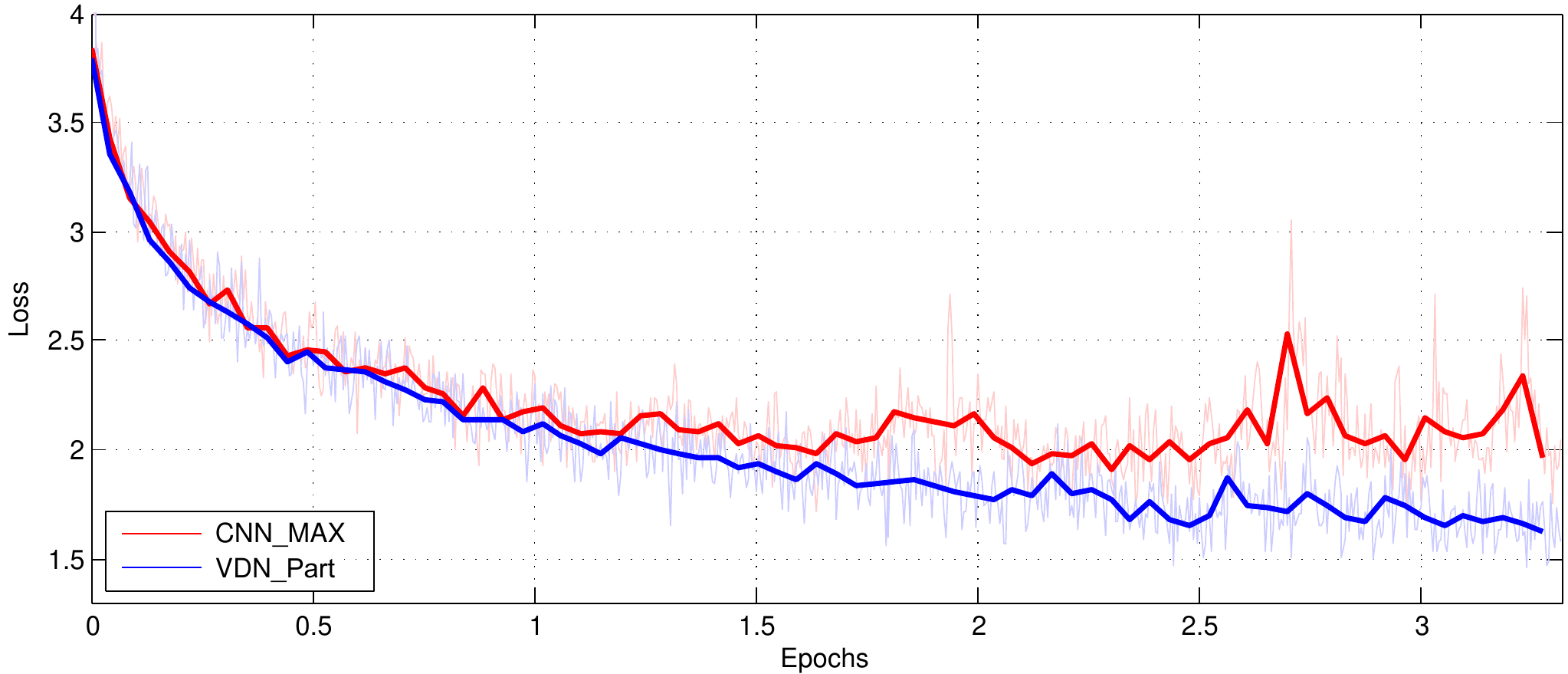}
\caption{Comparison on loss curves during the training.}
\label{lossCurve}
\end{figure}

The speed of loss convergence during training is an important criterion for the efficiency of the proposed network architecture. A higher rate of loss convergence represents less time and calculating consuming, which is crucial for practical applications. In this experiment, we aim to illustrate the validity of Score Generation Unit on improving the speed of loss convergence. To visualize the improvement, we add another Softmax loss to monitor the output vector of $inception\_5a/output$ layer, from which features are extracted and aggregated with score vectors. In this experiment, we use the ShapeNet Core55 perturbed dataset for training. Besides, one-quarter of each image is occluded by a black area to make the dataset more challenging. During the training, the loss curves are recorded for comparison. We compare our method with a baseline method, "CNN\_MAX", which adopts the same setting mentioned above. Fig. \ref{lossCurve} shows the curves of them.

As can be seen, loss of CNN with Score Generation Unit drops down stably, while the loss of "CNN\_MAX" suffers from the large and noisy dataset, which exactly confirms the capability of Score Generation Unit in efficiency improvement. The reason for this improvement lies in the interaction between CNN and Score Generation Unit. As the score vectors emphasize certain parts of the image feature, it influences the backward propagation process and brings a positive effect to the parameters of CNN1, improving the efficiency of extracting image features. This interaction proves that Score Generation Unit benefits the training in an all-around way, as it not only helps produce a robust shape feature but also enables the network to be quickly aware of positive information of each view.

\subsection{Limitations and Failure Cases}
Our View Discerning Network has two limitations.
First, the Part-wise Discerning Network takes an even division of an image into $49$ blocks,
which is oblivious to the specific shape in the image.
As such, important shape parts may be decomposed into several parts in different blocks, resulting in sub-optimal recognition accuracy.
Second, the Score Generation Unit directly takes raw images as input, leading to high computational cost.
Since the main network is pre-trained on ImageNet, it could extract valuable feature benefiting the scoring process.
Such information is, however, not utilized by the Score Generation Unit which is independent of the main network.
This separate architecture might be limiting the performance on score map generation.

\begin{figure}[!h]
\centering
\includegraphics[width = 3.3in]{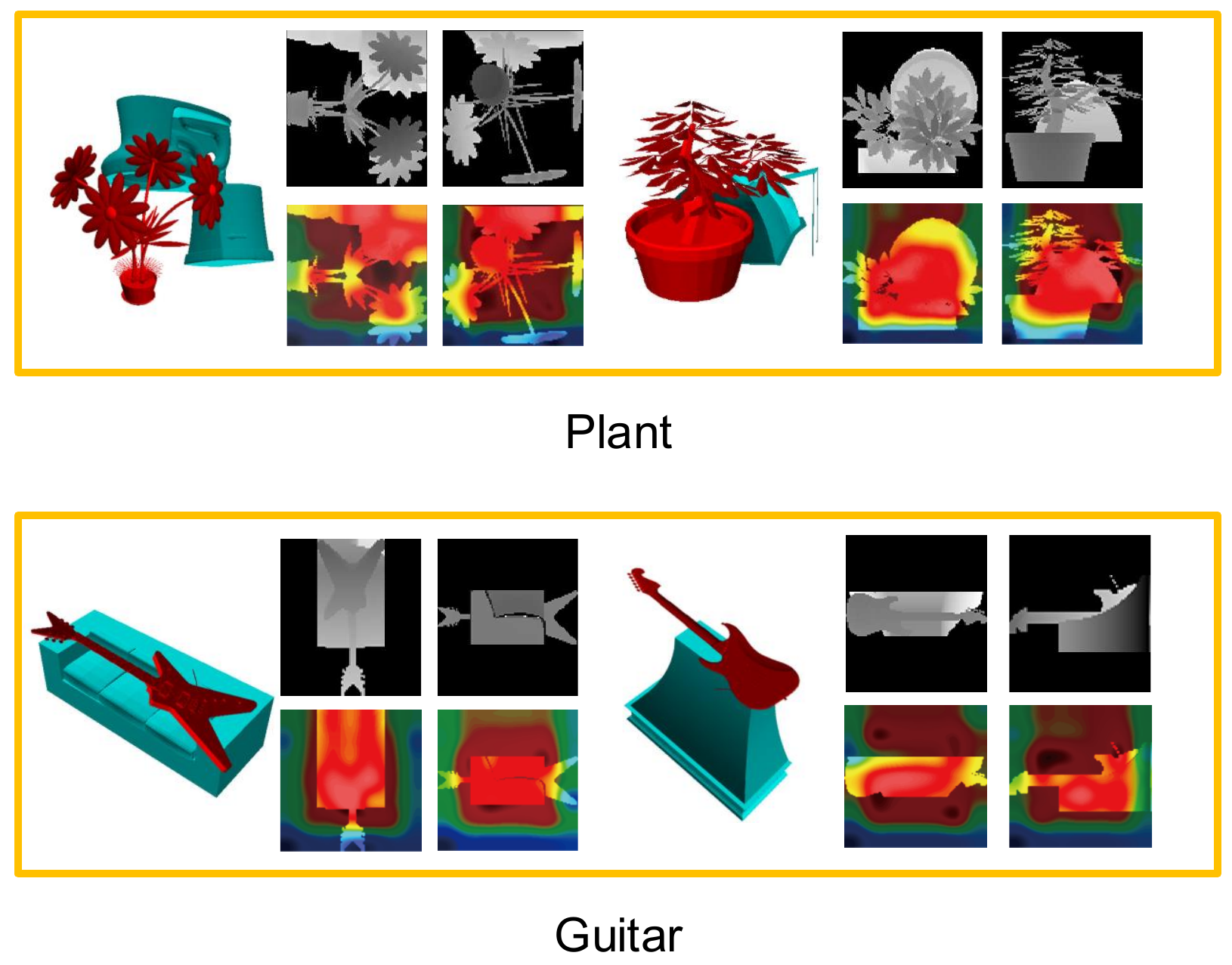}
\caption{Two typical failure cases. Top: Highly complex plant models, under occlusion, cannot be recognized correctly. Bottom: Our method fails to discern good views for the guitar models being put on larger models,
due to the confusion of foreground and background.}
\label{limit}
\end{figure}

Fig.~\ref{limit} shows two failure cases from occluded ModelNet40.
In the first example (top), the high complexity of plant models makes it hard for the Score Generation Unit to judge the quality of images correctly, due to the content-oblivious splitting of scoring blocks.
This is made even more severe when occlusion is added.
The classification accuracy for plant models is generally low, which is about $45\%$.
Please refer to the supplementary material for the statistics of per-category classification accuracy over the occluded ModelNet dataset.
The second example (bottom) shows two guitar models being put on larger models.
Our method fails to discern good views in this case, due to the confusion of foreground and background.

\section{Conclusion}
In this paper, we propose a deep neural network (View Discerning Network) to deal with the 3D shape recognition task. We pay attention to the quality of the input images and come out with a Score Generation Unit to make a judgment on the images. This unit produces a score vector for each input image. Then the network aggregates the image features with the scores to generate a weighted shape feature. In this case, images with good quality have a more positive influence on the shape representation. Meanwhile, the network can be less affected by poor images. Our experiments show that the View Discerning Network performs better than the state-of-the-art methods on ModelNet and ShapeNet core55. However, as some limitations exist in our method, the network needs further modification in the future work. First, the score maps can be constructed based on the feature points of a certain shape, rather than simply divides the image into uniform blocks. Second, the Score Generation Unit may share certain layers with the main network for better efficiency. Besides, the scores can be more reliable with the assist of the fine-tuned parameters.

\section{Acknowledgments}
This work is supported by the National Natural Science Foundation of China (No. 61472023, 61532003, 61572507,
and 61622212) and Beijing Municipal Natural Science Foundation (No. 4182034).

\begin{IEEEbiography}[{\includegraphics[width=1in,height=1.25in,clip,keepaspectratio]{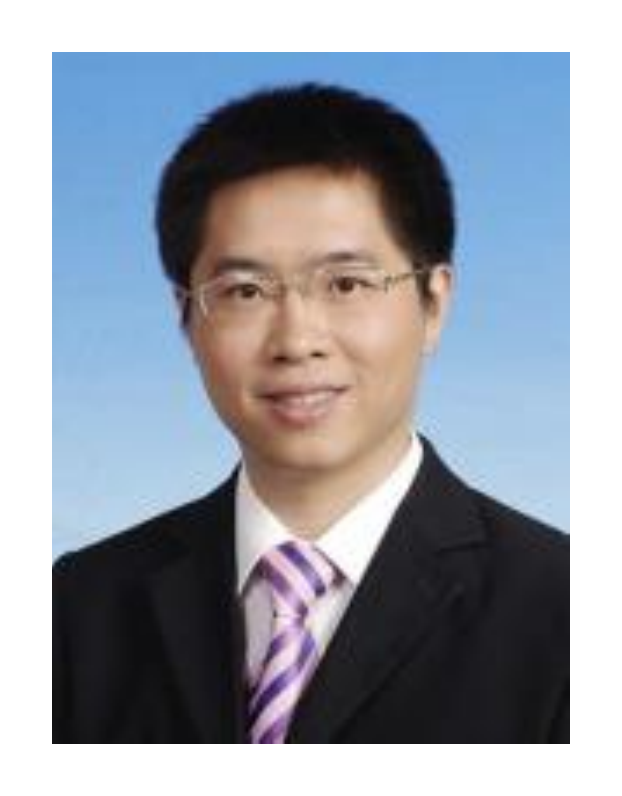}}]{Biao Leng}
received the bachelor’s degree from the School of Computer Science and Technology, National University of Defense Technology, Changsha, China, in 2004, and the Ph.D. degree from the Department of Computer Science and Technology, Tsinghua University, Beijing, China, in 2009. He is currently an Associate Professor with the School of Computer Science and Engineering, Beihang University, Beijing, China. His current research interests include 3D model retrieval, image processing, pattern recognition, and data mining.
\end{IEEEbiography}

\begin{IEEEbiography}[{\includegraphics[width=1in,height=1.25in,clip,keepaspectratio]{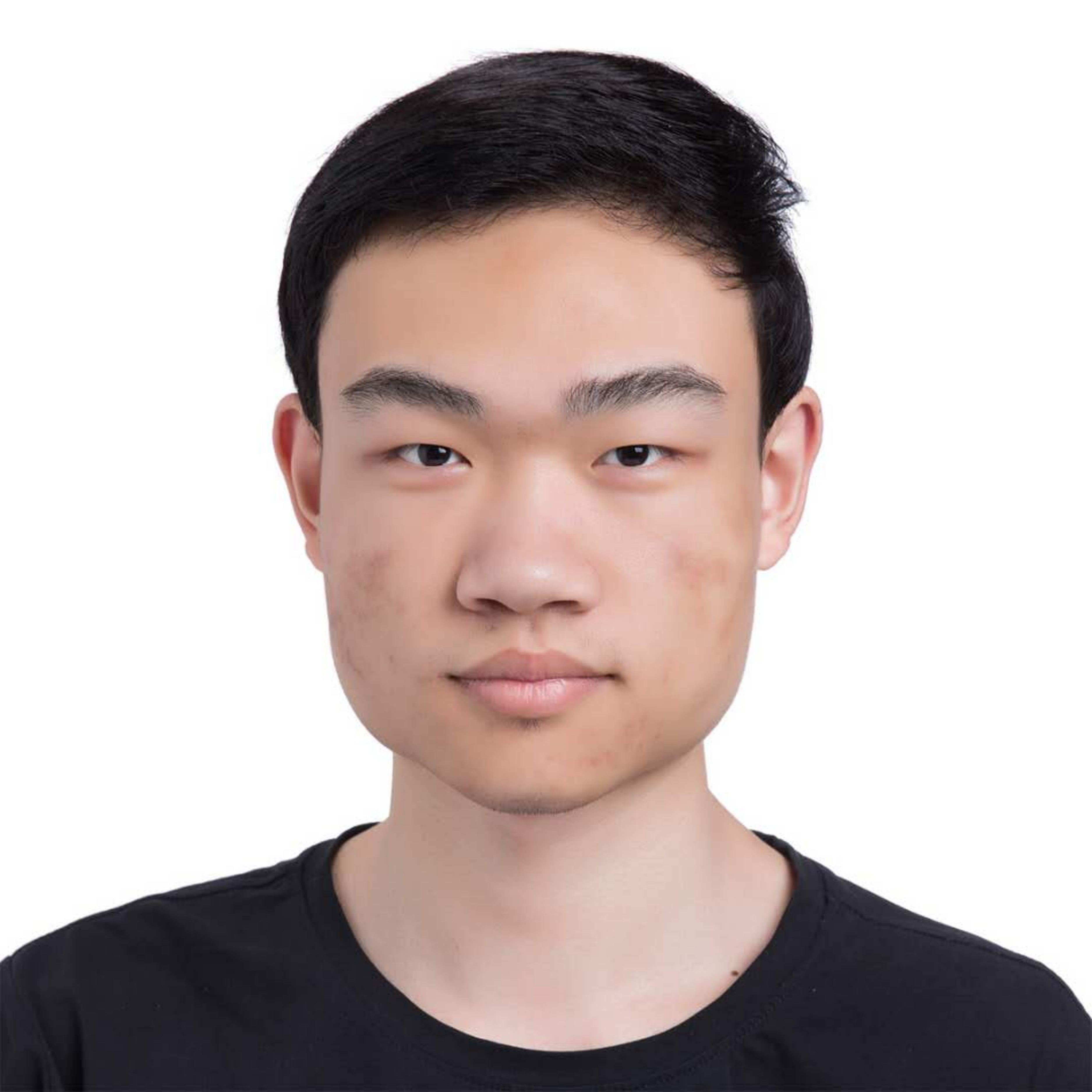}}]{Cheng Zhang}
received bachelor's degree in computer science and technology from Beihang University, Beijing, China, in 2018. He is currently pursuing the master's degree in School of Computer Science, Carnegie Mellon University, PA, USA. His research interests include 3D model recognition, computer vision and machine learning.
\end{IEEEbiography}

\begin{IEEEbiography}[{\includegraphics[width=1in,height=1.25in,clip,keepaspectratio]{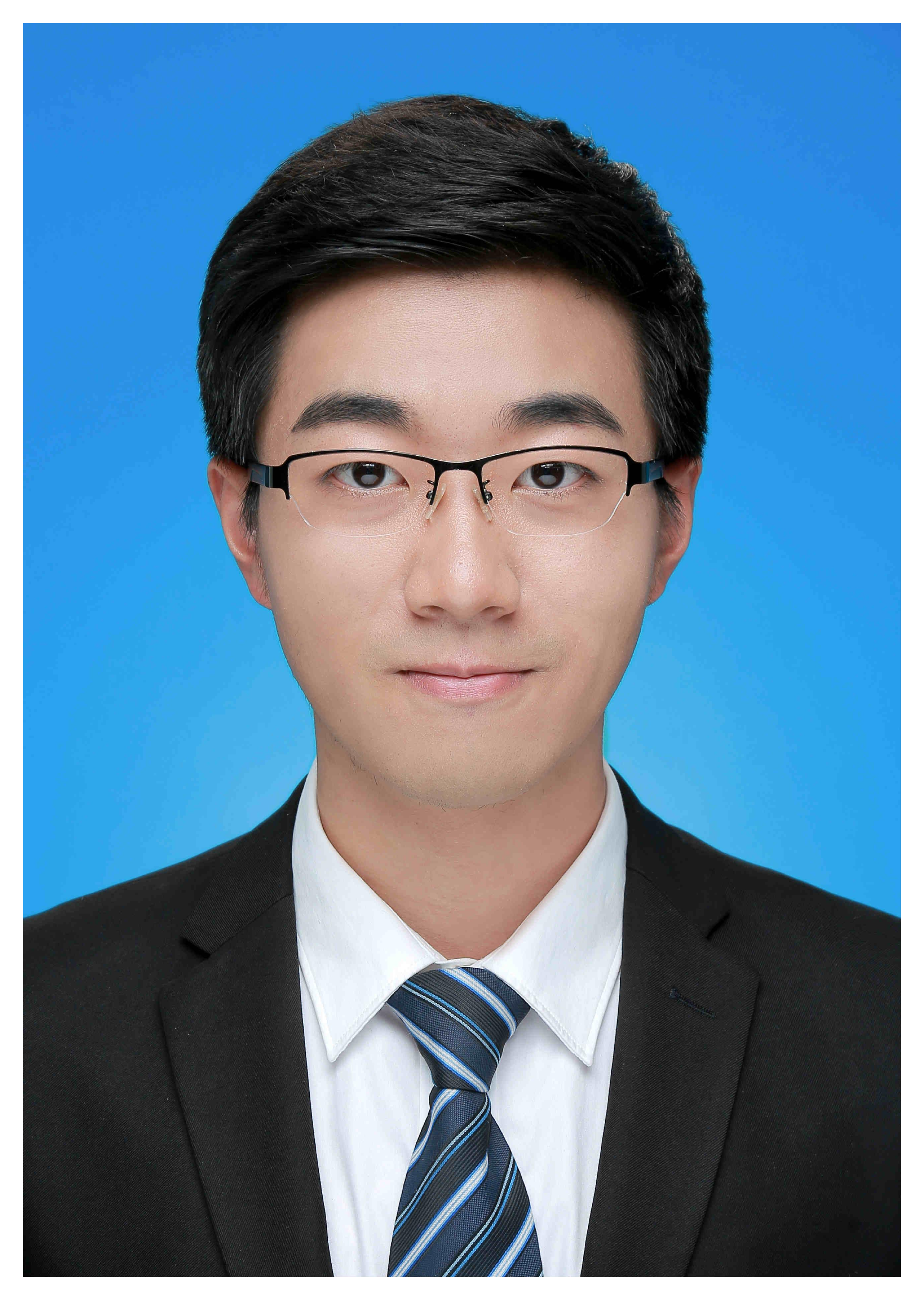}}]{Xiaochen Zhou}
received the bachelor's degree in Computer Science and Engineering from Beihang University, Beijing, China, in 2018. He is currently pursuing the master's degree in Washington University in St. Louis, School of Engineering and Applied Science, St. Louis, MO, USA. His research interests include 3D model retrieval, computer vision and machine learning.
\end{IEEEbiography}

\begin{IEEEbiography}[{\includegraphics[width=1in,height=1.25in,clip,keepaspectratio]{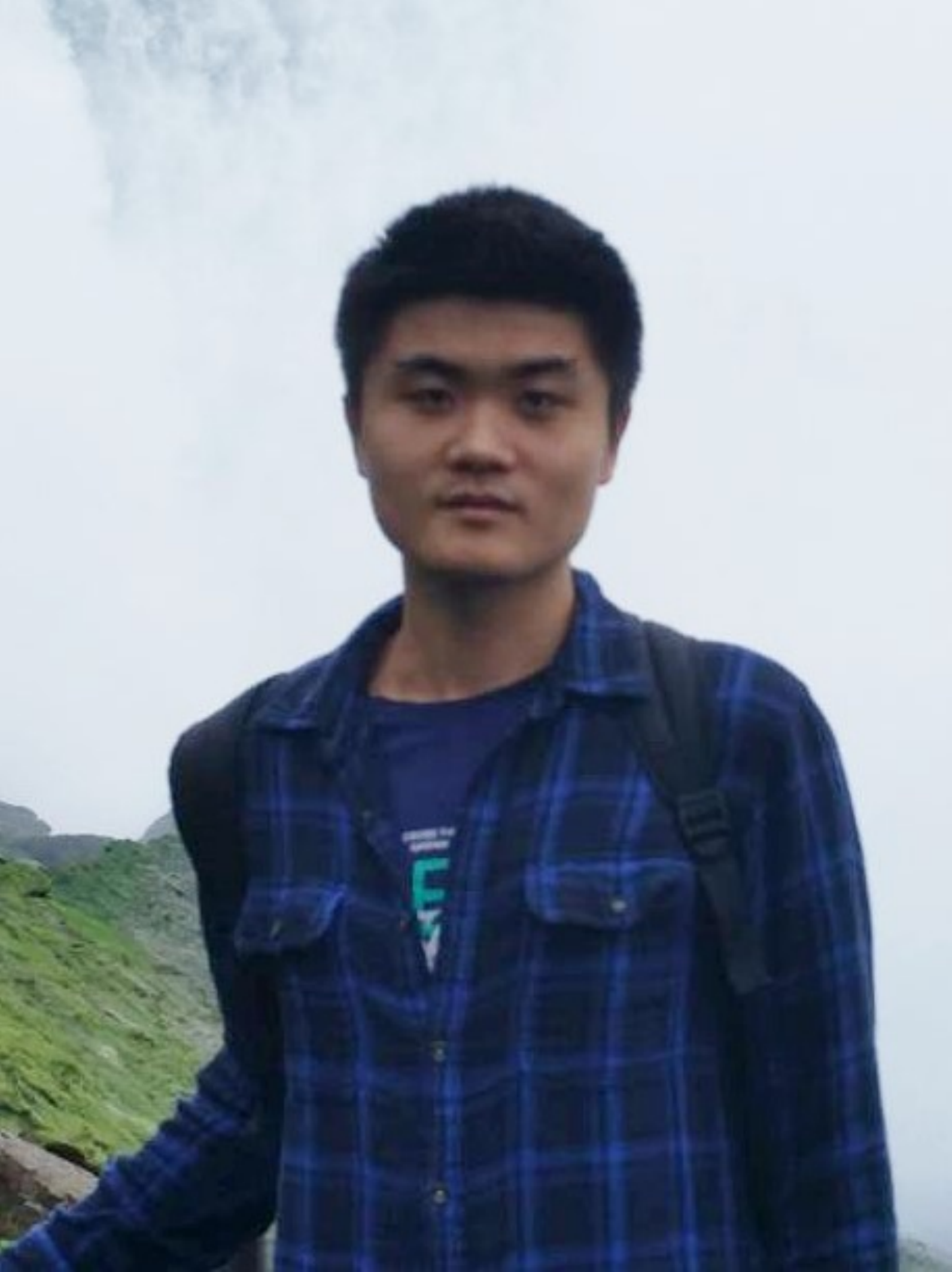}}]{Cheng Xu}
received the bachelor's degree in Computer Science and Engineering from Beihang University, Beijing, China, in 2017. He is currently pursuing the master's degree in the School of Computer Science and Engineering in Beihang University. His research is mainly on 3D shape recognition and shape understanding.
\end{IEEEbiography}

\begin{IEEEbiography}[{\includegraphics[width=1in,height=1.25in,clip,keepaspectratio]{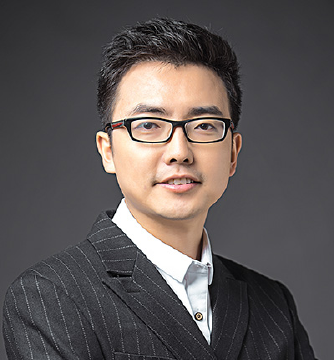}}]{Kai Xu}
is an Associate Professor at the School of Computer, National University of Defense Technology, where he received his Ph.D. in 2011. He conducted visiting research at Simon Fraser University (2008-2010) and Princeton University (2017-2018). His research interests include geometry processing and geometric modeling, especially on data-driven approaches to the problems in those directions, as well as 3D vision and its robotic applications. He is currently serving on the editorial board of Computer Graphics Forum, Computers \& Graphics, and The Visual Computer. He also served as paper co-chair of CAD/Graphics 2017 and ICVRV 2017, as well as PC member for several prestigious conferences including SIGGRAPH Asia, SGP, PG, GMP, etc. His research work can be found in his personal website: www.kevinkaixu.net.
\end{IEEEbiography}

\end{document}